\documentclass[submission,copyright,creativecommons]{eptcs}
\providecommand{\event}{FMAS 2023} 

\usepackage{todonotes}

\usepackage{rwthcolors}
\usepackage{amsfonts,amsmath,amsthm,amssymb}
\usepackage{mathtools}

\usepackage[toc,page]{appendix}
\usepackage{booktabs}
\usepackage{enumitem}
\usepackage{floatrow}
\usepackage{hyperref}
\usepackage{iftex}
\usepackage{lipsum}
\usepackage{multicol}
\usepackage{multirow}
\usepackage{siunitx}
\usepackage{subcaption}
\usepackage{tabularx}
\usepackage{longtable}
\usepackage{textcomp, gensymb}
\usepackage{wrapfig}

\usepackage{algorithm}
\usepackage{algorithmicx}
\usepackage[noend]{algpseudocode}

\usepackage{graphicx,wrapfig}
\usepackage{tikz}
\usepackage{tkz-base}
\usepackage{tkz-euclide}
\usetikzlibrary{overlay-beamer-styles,fit}
\newcommand\addvmargin[1]{
  \node[fit=(current bounding box),inner ysep=#1,inner xsep=0]{};
}

\ifpdf
  \usepackage[strings]{underscore}         
  \usepackage[T1]{fontenc}        
\else
  \usepackage{breakurl}           
\fi

\theoremstyle{definition}

\newtheorem{proposition}{Proposition}[section]

\newtheorem{definition}{Definition}[section]

\DeclarePairedDelimiter\abs{\lvert}{\rvert}%
\DeclarePairedDelimiter\norm{\lVert}{\rVert}%

\makeatletter
\let\oldabs\abs
\def\abs{\@ifstar{\oldabs}{\oldabs*}}
\let\oldnorm\norm
\def\norm{\@ifstar{\oldnorm}{\oldnorm*}}
\makeatother

\newcommand{\vectorS}[1]{\mathbf{#1}}   
\newcommand{\matrixS}[1]{\mathbf{#1}}   

\newcommand{\layerState}[1]{\vectorS{x}_{#1}}   
\newcommand{\layer}[1]{l_{#1}}                  
\newcommand{\layerSize}[1]{\langle #1 \rangle}    
\newcommand{\Weights}[1]{\matrixS{W}_{#1}}      
\newcommand{\bias}[1]{b_{#1}}  
\newcommand{\biases}[1]{\vectorS{b}_{#1}}       
\newcommand{\actFun}[3]{\vectorS{act}_{#1}^{#2}(#3)}
\newcommand{\actFunOA}[3]{\mathit{\overline{\vectorS{act}}}_{#1}^{#2}(#3)}

\newcommand{\R}{\mathbb{R}}
\newcommand{\N}{\mathbb{N}}

\newcommand{\sSet}[0]{\theta}
\newcommand{\sCenter}[1]{\vectorS{c}_{#1}}              
\newcommand{\sGenerator}[2]{\vectorS{v}^{(#1)}_{#2}}    
\newcommand{\sGeneratorM}[1]{\matrixS{V}}          
\newcommand{\sPredicate}[0]{P}
\newcommand{\sSum}[1]{\sum_{j=1}^{m}{#1}}               
\newcommand{\sStarVars}[1]{\vectorS{x}_{#1}}            
\newcommand{\sPolyVars}[1]{\vectorS{\alpha}_{#1}}       
\newcommand{\sPolyVarsVec}[0]{\vectorS{\alpha}}
\newcommand{\sPolyVarsAsVec}[0]{[\sPolyVars{1}, \hdots, \sPolyVars{m}]^{\intercal}}
\newcommand{\sPredWVars}[0]{(\sPolyVars{1}, \hdots, \sPolyVars{m})^T\in\sPredicate}
\newcommand{\sPredWVarVec}[0]{\sPolyVarsVec\in\sPredicate}

\newcommand{\reals}[0]{\mathbb{R}}                    
\newcommand{\reachSet}[1]{\mathcal{R}_{#1}}   
\newcommand{\polytope}[0]{\mathcal{P}}
\newcommand{\halfspace}[0]{\mathcal{H}}

\newcommand{\nullVec}[0]{[0, 0, \hdots, 0]^{\intercal}}
\newcommand{\stdBasisVec}[1]{\vectorS{e}_{#1}}
\newcommand{\stdBasis}[0]{\{ \stdBasisVec{1},  \hdots, \stdBasisVec{n} \}}
\newcommand{\suchthat}[0]{\; \text{such} \; \text{that} \;}
\newcommand{\suchthatshort}[0]{\; \text{s.} \; \text{t.} \;}
\newcommand{\setIf}[0]{\mid}                      
\newcommand{\tuple}[1]{\langle{#1}\rangle}        

\renewcommand{\triangleq}{=}

\algrenewcommand\algorithmicrequire{\textbf{Input:}}
\algrenewcommand\algorithmicensure{\textbf{Output:}}
\algnewcommand\algorithmicswitch{\textbf{switch}}
\algnewcommand\algorithmiccase{\textbf{case}}
\algnewcommand\algorithmicassert{\texttt{assert}}
\algnewcommand\Assert[1]{\State \algorithmicassert(#1)}%
\algdef{SE}[SWITCH]{Switch}{EndSwitch}[1]{\algorithmicswitch\ #1\ \algorithmicdo}{\algorithmicend\ \algorithmicswitch}%
\algdef{SE}[CASE]{Case}{EndCase}[1]{\algorithmiccase\ #1}{\algorithmicend\ \algorithmiccase}%
\algtext*{EndSwitch}%
\algtext*{EndCase}%

\title{Extending Neural Network Verification to a Larger Family of Piece-wise Linear Activation Functions}

\def\titlerunning{Extending Neural Network Verification to Piece-wise Linear Activations}

\author{
  László Antal
  \institute{RWTH Aachen University \\ Aachen, Germany}
  \email{antal@cs.rwth-aachen.de}
  \and  
  Hana Masara
  \institute{RWTH Aachen University \\ Aachen, Germany}
  \email{hana.masara@rwth-aachen.de}
  \and
  Erika Ábrahám
  \institute{RWTH Aachen University \\ Aachen, Germany}
  \email{abraham@cs.rwth-aachen.de}
}

\def\authorrunning{L. Antal, H. Masara \& E. Ábrahám}

\begin{document}

\maketitle

\begin{abstract}

In this paper, we extend an available neural network verification technique to support a wider class of \textit{piece-wise linear} activation functions. Furthermore, we extend the algorithms, which provide in their original form exact respectively over-approximative results for bounded input sets represented as star sets, to allow also \textit{unbounded} input sets.
We implemented our algorithms and demonstrated their effectiveness in some case studies.



\end{abstract}

\section{Introduction}
\label{sec:introduction}



In the area of artificial intelligence, \emph{feed-forward neural networks (FNNs)} \cite{lecun2015deep} enjoy increasing popularity. FNNs can be trained to learn a function $f:\mathbb{R}^n\rightarrow\mathbb{R}^m$ from a set of input-output samples, and predict outputs also for previously unseen inputs. This way, FNNs can tackle problems that would otherwise require very complex solutions \cite{samek2021explaining}.

Nowadays, a wide range of applications use FNNs, such as autonomous vehicles \cite{kuutti2020survey}, speech- and object-recognition systems \cite{hinton2012deep,erhan2014scalable} or robot vision \cite{lee2015comparing}, just to mention a few. While FNNs are impressively effective, their reliability in safety-critical situations is still questionable \cite{cubuk2017intriguing,goodfellow2014explaining,huang2017safety}.
Hence, verification methods play an important role in providing guarantees about their behavior. In this work, we focus on the \emph{reachability problem} for FNNs, which is the problem of determining which output values (\emph{reachable set}) an FNN computes for inputs from a given set.

\paragraph*{Related work.}

The application of formal methods \cite{woodcock2009formal,clarke1996formal,hinchey1995applications,wing1990specifier} to verify the safety of neural networks began with \cite{10.1007/978-3-642-14295-6_24}. Since then, the verification of neural networks has gained significant attention from the formal methods research community \cite{tran2019parallelizable,xiang2018output,ehlers2017formal,cheng2017maximum,wang2018efficient,liu2021algorithms,fromherz2021fast,tran2021verification,boopathy2019cnn,henriksen2021deepsplit,katz2019marabou,sun2021probabilistic}.

Some of the available approaches encode the verification problems as logical formulae and use SMT-solvers for their solution \cite{katz2017reluplex,wu2022efficient,katz2019marabou,ehlers2017formal,wang2018efficient}. Another common technique is reachable set calculation \cite{huang2019reachnn,ruan2018reachability,lomuscio2017approach,tran2019parallelizable,xiang2018output,tran2021verification} using an abstract representation like star sets \cite{Tran2021} or symbolic intervals \cite{KernBueningSinz2022}. 

This paper builds on previous work \cite{bak2017serllswi,10.1007/978-3-030-30942-8_39,Tran2021}, which solves the reachability problem using \textit{star sets} to represent subsets of $\mathbb{R}^k$ for any $k\in\mathbb{N}$ with $k>0$, like sets of input and output values.
The authors present two methods, one with exact computations and one which over-approximates the reachable set.



\paragraph*{Contributions.} 

Our contributions in this paper are the following:

\begin{enumerate}[topsep=0pt,itemsep=-1ex,partopsep=1ex,parsep=1ex]

\item We extend the set of activation functions supported by \cite{Tran2021,10.1007/978-3-030-30942-8_39} to cover the piece-wise linear functions \emph{leaky ReLU}, \emph{hard tanh}, \emph{hard sigmoid} and the \emph{unit step}; while some of these functions have already been included in the respective algorithms, no complete formalizations were available, which we provide in this paper. Furthermore, we support more general, \emph{parameterized} versions of the aforementioned activation functions.
For each of the above, we present the reachability analysis algorithm using both the exact and the over-approximative methods.

\item While previous work was restricted to bounded input sets, we provide extensions to allow also \emph{unbounded} input sets.

\item Using the open-source library HyPro\footnote{Implementation available online at \url{https://github.com/hypro/hypro}. For reproducing the experimental results, please check the Case Studies/Neural Network Verification subsection of HyPro's GitHub page: see the \texttt{README.md} file.} \cite{schupp2017h} for the star-set representation, we developed a C++ \emph{implementation} of both the exact and the over-approximative analysis methods, covering all the above activation functions.
This includes also an extension of HyPro with an NNET parser to input FNN models in NNET file format.

\item We propose some novel benchmarks (thermostat and sonar classifier) with the aim of supporting the formal methods community. Using our implementation, we provide \emph{experimental evaluation} on the two proposed benchmarks and two other existing benchmarks discussing the results.

\end{enumerate}

\paragraph*{Outline.} 

The rest of this paper is structured as follows. We present in Section \ref{sec:preliminaries} the fundamentals of this work, including feedforward neural networks (FNN), star sets, and reachability analysis of FNN with the rectified linear unit (ReLU) activation function. Then, in Section \ref{sec:method}, we propose an exact and over-approximate analysis method for several other activation functions, considering both bounded and unbounded input sets. Afterwards, in Section \ref{sec:evaluation}, we present and evaluate experimental results on four different benchmarks. Finally, in Section \ref{sec:conclusion} we conclude the paper and discuss future work.

\section{Preliminaries}
\label{sec:preliminaries}


We use $\N$ to denote the set of all natural numbers including 0 and $\R$ for the reals, and consider elements from $\R^n$ (for any $n\in\N$) to be column vectors.

\subsection{Feedforward Neural Networks}
\label{subsec:neural-networks}

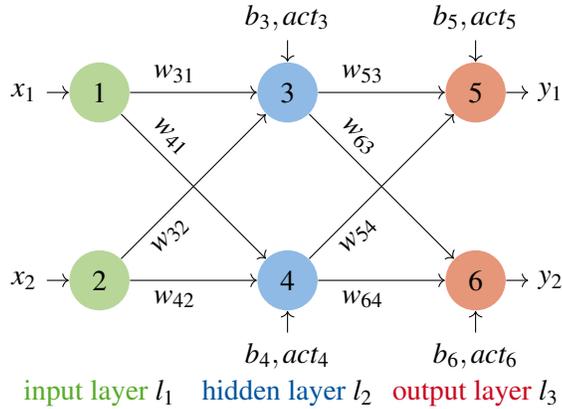
\begin{figure}
    \floatbox[{\capbeside\thisfloatsetup{capbesideposition={right,center},capbesidewidth=4.5cm}}]{figure}[\FBwidth]
             {\caption{Illustration of a feed-forward fully-connected neural network, consisting of one input layer (green), one hidden layer (blue), and one output layer (red).
               }\label{fig:neural-network}}
    {\begin{tikzpicture}[node distance=2.5cm]
    \node[circle, minimum size = 8mm,fill=green50] [] (I1) {$1$}; 
    \node[circle, minimum size = 8mm,fill=green50] [below of=I1] (I2) {$2$}; 
    
    \node[node distance=1cm] [left of=I1] (It1) {$x_1$}; 
    \node[node distance=1cm] [left of=I2] (It2) {$x_2$}; 
    
    \draw [->] (It1) -- (I1);
    \draw [->] (It2) -- (I2);
    
    \node[node distance=1.5cm] [below of=I2] (Ilt) {\textcolor{green100}{input layer} $l_1$};
        
    \node[circle, minimum size = 8mm,fill=blue50] [right of=I1] (H1) {$3$};
    \node[circle, minimum size = 8mm,fill=blue50] [below of=H1] (H2) {$4$};
    
    \node[node distance=1.5cm] [below of=H2] (Hlt) {\textcolor{blue100}{hidden layer} $l_2$};
    
    \draw [->] (I1) -- (H1) node [pos=0.35, above] {$w_{31}$};
    \draw [->] (I1) -- (H2) node [pos=0.25, rotate=-45, above] {$w_{41}$};
    \draw [->] (I2) -- (H1) node [pos=0.25, rotate=+45, below] {$w_{32}$};
    \draw [->] (I2) -- (H2) node [pos=0.35, below] {$w_{42}$};
    
    \node[node distance=1cm] [above of=H1] (b3) {$b_{3},\textit{act}_3$}; 
    \node[node distance=1cm] [below of=H2] (b4) {$b_{4},\textit{act}_4$}; 
    
    \draw [->] (b3) -- (H1);
    \draw [->] (b4) -- (H2);
    
    
    
    
    \node[circle, minimum size = 8mm,fill=red50] [right of=H1] (H3) {$5$};
    \node[circle, minimum size = 8mm,fill=red50] [right of=H2] (H4) {$6$};
    
    \draw [->] (H1) -- (H3) node [pos=0.35, above] {$w_{53}$};
    \draw [->] (H1) -- (H4) node [pos=0.25, rotate=-45, above] {$w_{63}$};
    \draw [->] (H2) -- (H3) node [pos=0.25, rotate=+45, below] {$w_{54}$};
    \draw [->] (H2) -- (H4) node [pos=0.35, below] {$w_{64}$};
    
    \node[node distance=1cm] [above of=H3] (b5) {$b_{5},\textit{act}_5$}; 
    \node[node distance=1cm] [below of=H4] (b6) {$b_{6},\textit{act}_6$}; 
    
    \draw [->] (b5) -- (H3);
    \draw [->] (b6) -- (H4);
    
    \node[node distance=1.5cm] [below of=H4] (H2lt) {\textcolor{red100}{output layer} $l_3$};
    
    
    
    
    \node[node distance=1cm] [right of=H3] (o1) {$y_1$}; 
    \node[node distance=1cm] [right of=H4] (o2) {$y_2$}; 
    
    \draw [->] (H3) -- (o1);
    \draw [->] (H4) -- (o2);
    
\end{tikzpicture}}
\end{figure}


A \textit{feedforward neural network (FNN)} \cite{kumar2019fnne,
  svozil1997intromutlifnn} is a directed weighted graph annotated with
some data.  It has a finite set of nodes called \emph{neurons}, which
are grouped into $k\in\N_{\geq 2}$ disjoint non-empty ordered sets
$\layer{1},\ldots,\layer{k}$ called \emph{layers}. We call $\layer{1}$
the \emph{input layer}, $\layer{k}$ the \emph{output layer}, while the
others are \emph{hidden layers}.  Let $\layerSize{i}$ denote the size $|\layer{i}|$ of layer $i=1,\ldots,k$. There is a directed edge from each
neuron $n$ in each non-output layer $\layer{i-1}$ to each neuron $n'$ in
the next layer $\layer{i}$, weighted by $w_{n',n}\in\R$; let
$\Weights{i}\in\R^{\layerSize{i}\times\layerSize{i-1}}$ be the matrix whose entry in row $r$ and column $c$ is the
weight of the edge from the $c$th neuron in layer $i-1$ to the $r$th
neuron in layer $i$.
In addition, each neuron $n$ in each non-input
layer is annotated with a \emph{bias} $\bias{n}\in \R$ and an
\emph{activation function} $\textit{act}_{n}:\R\rightarrow\R$; for
layer $i$ with neurons $\layer{i}=\{n_1,\ldots,n_{\layerSize{i}}\}$,
let $\biases{i}=(\bias{n_1},\ldots,\bias{n_{\layerSize{i}}})^T$ and
$\vectorS{act}_{i}:\R^{\layerSize{i}}\rightarrow\R^{\layerSize{i}}$
with $\vectorS{act}_i(\vectorS{y})=(\textit{act}_{n_1}(y_1),\ldots,\textit{act}_{n_{\layerSize{i}}}(y_{\layerSize{i}}))^T$
for any input $\vectorS{y}=(y_1,\ldots,y_{\layerSize{i}})^T\in\R^{\layerSize{i}}$.
A frequently used activation function is the \emph{Rectified Linear Unit (ReLU)}, defined as $ReLU(x) = max(0, x)$ for $x\in\R$.
An example FNN is shown in \autoref{fig:neural-network}.

For an \emph{input} $\vectorS{x}_1=(x_1,\ldots,x_{\layerSize{1}})\in \R^{\layerSize{1}}$, the \emph{state} $\layerState{i}$ of each non-input layer $\layer{i}$ is defined recursively as
\begin{equation}
  \layerState{i} = \actFun{i}{}{\Weights{i}\, \layerState{i-1} + \biases{i}} \ .
\end{equation}
\noindent Thus an FNN can be seen as a function
$ f : \R^{\layerSize{1}} \to \R^{\layerSize{k}} $, assigning to each input the output layer's state, which we call the \emph{output}.
%
%
%
%
For a given FNN and a set
$\reachSet{1}$ of possible inputs, the \emph{FNN reachability problem} is the problem to compute all possible states for each of the layers $1<i\leq k$ \cite{10.1007/978-3-030-30942-8_39}:
\begin{eqnarray}
    \label{eqn:reach-set}
    \reachSet{i} & = & \{ \,\vectorS{act}_i(\Weights{i} \sStarVars{i-1} + \biases{i})\setIf \sStarVars{i-1} \in \reachSet{i-1} \,\} \ .
\end{eqnarray}
Solving the FNN reachability problem allows us to check properties of interest, e.g. safety properties (whether the output set is disjoint from a set of unsafe outputs) or stability (whether the distance between possible outputs is below a threshold for a given input set).

In this work, as input set we consider convex polyhedra $\reachSet{1} = \{ \sStarVars{} \in \R^{\layerSize{1}} \setIf \matrixS{A} \sStarVars{} \leq \vectorS{c}\, \}$ for some $m\in\N_{\geq 1}$, $\matrixS{A}\in\R^{m\times \layerSize{1}}$ and column vector $\vectorS{c}\in\R^{m}$.




\subsection{Stars}
\label{subsec:star-sets}

To compute $\reachSet{i}$ via \autoref{eqn:reach-set}, the two main operations that need to be applied on state sets are the activation function $\vectorS{act}_i$ and \textit{affine transformations} using the weights $\Weights{i}$ and biases $\biases{i}$ of the layer $i$. For implementing these calculations efficiently, different state set representations have been proposed \cite{Schupp2019StateSR}. Under these, star sets (or short stars) turned out to be exceptionally good candidates, for their efficient handling of affine transformations and half-space intersections (see Propositions \ref{prop:aff-trans}, \ref{prop:halfspace-intersect} and \ref{prop:emptiness}).

For any $n,m\in\N$, an $(n,m)$-dimensional \emph{star} is a tuple $\sSet=\tuple{\sCenter{}, \sGeneratorM{}, \sPredicate{}}$ of (i) a \emph{center} $\sCenter{} \in \R^n$, (ii) a \emph{generator matrix} $\sGeneratorM{}\in\R^{n\times m}$ whose columns $\sGenerator{1}{}, \ldots, \sGenerator{m}{}\in \R^n$ are called the \emph{basis vectors} or \emph{generators} and (iii) a \emph{predicate} $\sPredicate\subseteq \R^m$. The star $\sSet$ \emph{represents} the set
$
[ \sSet ] = \{\, \sCenter{} + \sSum{(\sPolyVars{j} \sGenerator{j}{})}  \setIf  \sPredWVars \,\}
$.

As in \cite{10.1007/978-3-030-30942-8_39}, we restrict $\sPredicate$ to be a convex polyhedron
$\sPredicate = \{ \sPolyVarsVec \in \R^{m} \setIf \matrixS{C} \sPolyVarsVec \leq \vectorS{d}\, \}$ for some $p\in\N$, $\matrixS{C} \in \R^{p \times m}$ and $\vectorS{d} \in \R^p$.
The following star properties, whose proofs are included in Appendix \ref{subsec:formal-proofs}, will be used to solve the FNN reachability problem.

\begin{proposition}[Convex polyhedra]
    For any $m,p\in\N$, $\matrixS{C} \in \R^{p \times m}$ and $\vectorS{d} \in \R^p$, the convex polyhedron $\polytope \triangleq \{ \sStarVars{}\in \R^{m} \setIf \matrixS{C} \sStarVars{} \leq \vectorS{d} \}$ can be represented by a star.

    \label{prop:conv-poly}
\end{proposition}

\begin{proposition}[Affine transformation]
  Assume an $(n,m)$-dimensional star $\sSet \triangleq \tuple{\sCenter{}, \sGeneratorM{}, \sPredicate}$ and let $\Weights{}\in\R^{k\times n}$ and $\biases{}\in\R^k$.
Then the affine transformation $\{ \Weights{} \sStarVars{} + \biases{} \setIf  \sStarVars{} \in [\sSet] \}$ of $[\sSet]$ is represented by
  $\bar{\sSet} = \tuple{\bar{\sCenter{}}, \bar{\sGeneratorM{}}, P}$ with 
  $\bar{\sCenter{}} = \Weights{} \sCenter{} + \biases{}$ and 
  $\bar{\sGeneratorM{}}\in\R^{k\times m}$ with columns $\Weights{} \sGenerator{1}{},  \hdots, \Weights{} \sGenerator{m}{}$.

    \label{prop:aff-trans}
\end{proposition}

\begin{proposition}[Intersection with halfspace]
    Assume an $(n,m)$-dimensional star $\sSet \triangleq \tuple{\sCenter{}, \sGeneratorM{}, \sPredicate}$ and a half-space $\halfspace \triangleq \{ \sStarVars{}\in\R^n \setIf \vectorS{h}^T \sStarVars{} \leq g \}$ with some $\vectorS{h}\in\R^{n}$ and $g\in\R$. Then the intersection $[\sSet]\cap\halfspace$ is represented by the star $\bar{\sSet}=\tuple{\sCenter{}, \sGeneratorM{}, \sPredicate \cap {\sPredicate}'}$ with 
    $
    {\sPredicate}' = \{\sPolyVarsVec\in\R^m\,|\, (\vectorS{h}^T  \sGeneratorM{m}) \sPolyVarsVec \leq g - \vectorS{h}^T \sCenter{} \}
    $.
\label{prop:halfspace-intersect}
\end{proposition}

\begin{proposition}[Emptiness check]
    A star $\sSet \triangleq \tuple{\sCenter{}, \sGeneratorM{}, \sPredicate}$ is empty if and only if $\sPredicate$ is empty. 

\label{prop:emptiness}
\end{proposition}

\begin{proposition}[Bounding box]
  Assume an $(n,m)$-dimensional star $\sSet \triangleq \tuple{\sCenter{}, \sGeneratorM{}, \sPredicate}$ with $\sCenter{}=(\sCenter{1},\ldots,\sCenter{n})^T$, and let $\sGeneratorM{}_{(i)}$ be the $i^{th}$ row of $\sGeneratorM{m}$. Let furthermore $B=\{(x_1,\ldots,x_n)^T\in\R^n\,|\,\bigwedge_{i=1}^n \textit{lb}_i\leq x_i\leq \textit{ub}_i\}$ with
        $\textit{lb}_i = \sCenter{i} + \underset{\sPredWVarVec}{\text{min}} \; \sGeneratorM{}_{(i)} \sPolyVarsVec $ 
and
        $\textit{ub}_i = \sCenter{i} + \underset{\sPredWVarVec}{\text{max}} \; \sGeneratorM{}_{(i)} \sPolyVarsVec $ for $i=1,\ldots,n$. Then $[\sSet]\subseteq B$.


\label{prop:bounding}
\end{proposition}



\subsection{Reachability Analysis for FNNs with ReLUs}
\label{subsec:reachability-analysis}

Next, we present two algorithms proposed in \cite{10.1007/978-3-030-30942-8_39} to solve the reachability problem for FNNs with the ReLU activation function for bounded polyhedral input sets. The first algorithm is exact and thus complete, whereas the second algorithm over-approximates reachability. We note that alongside ReLU, \cite{Tran2021} includes some other activation functions but no complete formalizations were available. In Section \ref{sec:method}, we will extend these algorithms to support further and more general piece-wise linear activation functions and unbounded input sets.

\subsubsection*{Exact Analysis}
\label{subsubsec:exact-analysis}

The exact algorithm first constructs a star from the input set which is required to be a polyhedron (see Proposition \ref{prop:conv-poly}). Then, correspondingly to \autoref{eqn:reach-set} it propagates the star through the network, layer-by-layer, until we get the output set $\reachSet{k}$. This propagation involves two main operations.

\noindent (1) For each non-input layer $i$ and each star representing possible states of the previous layer, to compute the reachable states of layer $i$, we first apply an affine transformation on the star, using the weight matrix $\Weights{i}$ and the bias vector $\biases{i}$. Thus, from a star $\sSet = \tuple{\sCenter{}, \sGeneratorM{}, \sPredicate}$ we obtain a new star $\sSet' = \tuple{\sCenter{}', \sGeneratorM{}', \sPredicate}$ with $\sCenter{}'=\Weights{i} \sCenter{} + \biases{i}$ and $\sGeneratorM{}'= \Weights{i} \sGeneratorM{}$ (see Proposition \ref{prop:aff-trans}). Note that during the affine transformation the predicate does not change.

\noindent (2) Then the non-linear activation function is applied on the intermediate star $\sSet'$ dimension-wise to represent $\reachSet{i} = \actFun{n_{\layerSize{i}}}{}{\hdots \, \actFun{n_1}{}{[\sSet']} \, \hdots}$,
where, $n_1,\ldots,n_{\layerSize{i}}$ are the neurons in layer $i$.
%
Since we consider the ReLU activation function, the $\actFun{n_j}{}{\cdot}$ operation at neuron $n_j$ is defined as $ReLU(x_j) = max(0, x_j)$; instead of $\actFun{n_j}{}{\cdot}$ we also write $\actFun{j}{\text{R}}{\cdot}$ to denote that the ReLU function is applied in dimension $j$ (i.e. at the $j$th neuron of a layer). To compute $\actFun{j}{\text{R}}{\sSet}$ for a star $\sSet = \tuple{\sCenter{}, \sGeneratorM{}, \sPredicate}$, the star $\sSet$ is decomposed into two stars  $\sSet_1 = \tuple{\sCenter{}, \sGeneratorM{}, \sPredicate_1}$ and $\sSet_2 = \tuple{\sCenter{}, \sGeneratorM{}, \sPredicate_2}$ such that  $[\sSet_1] = [\sSet] \cap \{(x_1,\ldots,x_n)\in\R^n\,|\,x_j < 0\}$ and $[\sSet_2] = [\sSet] \cap \{(x_1,\ldots,x_n)\in\R^n\,|\,x_j \geq 0\}$ (see Proposition \ref{prop:halfspace-intersect}).
On the negative branch, i.e., when $x_j < 0$, the ReLU function sets the corresponding values to zero. Thus all the resulting elements of the star $\sSet_1$ should have the value zero in dimension $j$. It affects the star as a projection to $0$ in dimension $j$. We can obtain this result by applying the mapping matrix $\matrixS{M} = [\vectorS{e}_1, \vectorS{e}_2, \hdots, \vectorS{e}_{j-1}, \vectorS{0}, \vectorS{e}_{j+1}, \hdots \vectorS{e}_n]$ on $\sSet_1$, where $\vectorS{e}_i\in\R^n$ is the $i$th $n$-dimensional unit vector (with $1$ at position $i$ and $0$s otherwise). On the positive branch $x_j \geq 0$, the ReLU function does not change the set elements of $\sSet_2$.
Thus, the application of ReLU results in the union of two stars $\actFun{j}{\text{R}}{\sSet} = \tuple{\matrixS{M}\sCenter{}, \matrixS{M}\sGeneratorM{}, \sPredicate_1} \cup \tuple{\sCenter{}, \sGeneratorM{}, \sPredicate_2}$. Note that if the values in $[\sSet]$ in the given dimension $j$ are purely positive or purely negative, then the result of $\actFun{j}{\text{R}}{\sSet}$ is just a single star.

\subsubsection*{Over-approximate Analysis}
\label{subsubsec:overapprox-analysis}

While the exact algorithm is complete, it suffers from scalability issues since the number of stars grows during the analysis \textit{exponentially} with the number of neurons. To tackle this problem, one solution is to side-step to over-approximative computations, which makes the analysis more \textit{scalable}, however, it sacrifices the \textit{completeness} of the method.

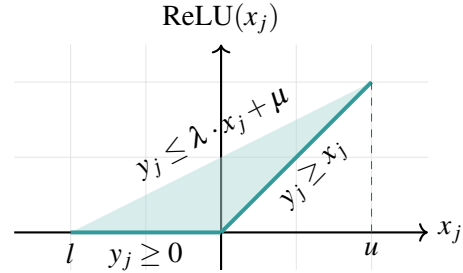
\begin{wrapfigure}[13]{r}{7.05cm}
  \centering
  \begin{tikzpicture}
    \draw[gray!20,step=1] (-2.75,-0.5) grid (2.75,2.5);
    
    \draw[thick,->] (-2.75, 0) -- (2.75, 0) node[right] {$x_j$};
    \draw[thick,->] (0, -0.5) -- (0, 2.5) node[above] {$\text{ReLU}(x_j)$};

    \fill[teal!30, opacity=0.5] (-2, 0) -- (2, 2) -- (0, 0);

    \draw (-2, 0) -- (0, 0) node[midway, below] {$y_j \geq 0$};
    \draw[thick, teal!80,line width=1.5pt] (-2, 0) -- (0, 0);
    \draw (0, 0) -- (2, 2) node[midway, below,rotate=46] {$y_j \geq x_j$};
    \draw[thick, teal!80,line width=1.5pt] (0, 0) -- (2, 2);
    \draw (-1, 0.5) node[above,rotate=27,xshift=11.8mm] {$y_j \leq \lambda \cdot x_j + \mu $};

    
    \draw[teal, dashed] (2,0) -- (2,2);
           
    \node[below] at (2, 0) {$u$};
    \node[below] at (-2, 0) {$l$};
    
\end{tikzpicture}

    
    
    




    

    

  \caption{Relaxation of the ReLU function with $\lambda = \frac{u}{u - l}$ and $ \mu = -\frac{l u}{u - l}$ \cite{DBLP:journals/pacmpl/SinghGPV19}. Dark lines represent the exact set, the light area shows the approximate set.
  }
  \label{fig:overapprox-relu}
\end{wrapfigure}

The over-approximate method from \cite{10.1007/978-3-030-30942-8_39}  also builds on \autoref{eqn:reach-set}, but the application of the activation functions is different:
the original $\actFun{j}{\text{R}}{\cdot}$ operation is replaced by an over-approximating $\actFunOA{j}{\text{R}}{\cdot}$ which produces only a single star as output as follows.
A new variable $\alpha_{m+1}$ and three more constraints are added to the predicate $\sPredicate$ of the star, with the purpose of capturing the over-approximation of the ReLU function at neuron $n_j$ (see \autoref{fig:overapprox-relu}).

The three new constraints are: $\alpha_{m+1} \geq 0$, $\alpha_{m+1} \geq x_j$, and $\alpha_{m+1} \leq \frac{u(x_j - l)}{u - l}$, where $l$ and $u$ are the lower and upper bounds, respectively, for variable $x_j$ in $[\sSet]$ (see Proposition \ref{prop:bounding}). Finally, since we want the variable $\alpha_{m+1}$ to hold the over-approximation of $x_j$, after introducing the new variable and constraints to the predicate, we need to update the center $\sCenter{}$ and basis $\sGeneratorM{}$ of the star $\sSet$ correspondingly. First, the old values of $x_j$ are projected out using the mapping matrix $\matrixS{M} = [\vectorS{e}_1, \vectorS{e}_2, \hdots, \vectorS{e}_{j-1}, \vectorS{0}, \vectorS{e}_{j+1}, \hdots \vectorS{e}_n]$. Then, a new generator vector $\vectorS{e}_j$ is added to the basis, to link $x_j$ to $\alpha_{m+1}$. 

Formally, for an $(n,m)$-dimensional star $\sSet=\tuple{\sCenter{}, \sGeneratorM{}, \sPredicate}$ we define $\actFunOA{j}{\text{R}}{\sSet} \triangleq \tuple{\bar{\sCenter{}}, \bar{\sGeneratorM{}}, \bar{\sPredicate}}$, where $\bar{\sCenter{}} = \matrixS{M} \sCenter{}$, $\bar{\sGeneratorM{}} = [\matrixS{M} \sGenerator{1}{}, \matrixS{M} \sGenerator{2}{}, \hdots, \matrixS{M} \sGenerator{m}{}, \vectorS{e}_{j}]$ and $\bar{\sPredicate} = \{(\alpha_1,\ldots,\alpha_{m+1})\in\R^{m+1}\,|\,(\alpha_1,\ldots,\alpha_m)\in\sPredicate \wedge \alpha_{m+1} \geq 0 \wedge \alpha_{m+1} \geq x_j \wedge \alpha_{m+1} \leq \frac{u(x_j - l)}{u - l}\}$.

In case $l \geq 0$ or $u \leq 0$, the introduction of a new variable is not necessary and we can proceed in a similar way as in the exact case, i.e., for positive domain we keep the set as it is, for negative domain we project out the variable $x_j$. Note that this over-approximation method is the least conservative that we can achieve using convex, linear constraints.

\section{FNN Reachability Analysis for Piece-wise Linear Activation Functions}
\label{sec:method}

Neural networks offer flexibility in choosing different activation
functions.  In this work, we present the extension of the reachability
analysis algorithm to implement the \emph{leaky rectified linear unit
(leaky ReLU)}, \emph{hard hyperbolic tangent (HardTanh)}, \emph{hard
sigmoid (HardSigmoid)}, and \emph{unit step} activation functions.
Below we define each of these functions and their application to a
given star $\sSet = \tuple{\sCenter{}, \sGeneratorM{}, \sPredicate}$.

\subsection{Unbounded Input Sets}
\label{subsec:unbounded}


During the analysis of an FNN, it may happen that one or more variables $x_j$ of a star  $\sSet$ become unbounded. That is, it has no lower bound (i.e., $l = -\infty$) or it has no upper bound (i.e., $u = \infty$). In the following, we present how to handle unbounded input sets as well.

Essentially, the exact reachability analysis of any piece-wise linear activation function presented in this paper does not change in case of unbounded input sets. The same steps are applied as per the exact analysis of bounded sets, i.e., (1) splitting the input set into multiple subsets based on the cases of the activation function, and (2) applying the corresponding transformations for each subset.

Conversely, in case of unbounded input, the over-approximate analysis does work differently, since the convex relaxations presented for bounded input need to be changed. In the rest of this paper, for each activation function, we show how the convex relaxations can be adjusted to achieve the \textit{tightest} possible relaxation in case of unbounded inputs. Note that we distinguish for each function three cases of unboundedness of a variable $x_j$, either it has no lower bound ($l = -\infty$ and $u \in \reals$), it has no upper bound ($l \in \reals$ and $ u = \infty$), or it has neither of the bounds ($l = -\infty$ and $ u = \infty$).

Our implementation currently does not support unbounded input sets, so the presented methods for unbounded inputs are only theoretical results. Furthermore, the evaluated benchmarks also do not utilize unbounded sets.

\subsection{Leaky ReLU Layer}
\label{subsec:leakyRelu-layer}

Due to the dead neuron problem \cite{datta2020surveyonactfunction,deadneuron2021} caused by the ReLU function, its alternative, the leaky ReLU function proposed by Mass et al.\cite{Maas2013RectifierNI}, is used in many applications.

\begin{definition}[Leaky ReLU \cite{xu2020reluplex}]
    The \emph{leaky ReLU} activation function with scaling parameter $\gamma \in (0,1)\subset\R$ is defined for each $x\in\R$ as 
    \begin{equation}
        LeakyReLU(x) = max(\gamma \cdot x, x) = \begin{cases}
            x & \textit{if } x > 0 \\
            \gamma \cdot x & \textit{otherwise}\ .
        \end{cases} 
    \end{equation}
\end{definition}

\subsubsection*{Exact Analysis}
\label{subsubsec:leakyRelu-exact-analysis}

The application of the leaky ReLU activation function is similar to the previously presented algorithm for the ReLU activation function, but they handle the negative inputs differently: While the ReLU function completely projects the input to zero, the leaky ReLU just scales the input down by $\gamma \in (0,1)$. Thus, the application $\actFun{j}{\text{L}}{\sSet}$ of leaky ReLU on a star $\sSet$ can be computed as follows. First we split the star $\sSet = \tuple{\sCenter{}, \sGeneratorM{}, \sPredicate}$ into two subsets $\sSet_1 = \tuple{\sCenter{}, \sGeneratorM{}, \sPredicate_1}$ and $\sSet_2 = \tuple{\sCenter{}, \sGeneratorM{}, \sPredicate_2}$ with negative resp. non-negative $x_j$-values. Then we apply the corresponding transformations for both subsets. As previously, in the case of the positive subset $\sSet_2$, no transformation is needed, since the leaky ReLU acts as an identity function for positive inputs. However, in case of the negative subset $\sSet_1$, we apply the scaling matrix $\matrixS{M} = [\vectorS{e}_1, \vectorS{e}_2, \hdots, \gamma \vectorS{e}_{j}, \hdots \vectorS{e}_{n-1}, \vectorS{e}_n]$. Thus, the final result of the $\actFun{j}{\text{L}}{\cdot}$ operation at neuron $n_j$ is the union of two stars: $\actFun{j}{\text{L}}{\sSet} = \tuple{\matrixS{M}\sCenter{}, \matrixS{M}\sGeneratorM{}, \sPredicate_1} \cup \tuple{\sCenter{}, \sGeneratorM{}, \sPredicate_2}$. The same observations apply here, that if the domain of a variable $x_j$ is only negative (i.e., $u \leq 0$) or only positive (i.e., $l \geq 0$), the final result of the $\actFun{j}{\text{L}}{\cdot}$ operation is a single star: either $\sSet_1 = \tuple{\matrixS{M}\sCenter{}, \matrixS{M}\sGeneratorM{}, \sPredicate_1}$ or $\sSet_2 = \tuple{\sCenter{}, \sGeneratorM{}, \sPredicate_2}$.

\subsubsection*{Over-approximate Analysis}
\label{subsubsec:leakyRelu-overapprox-analysis}

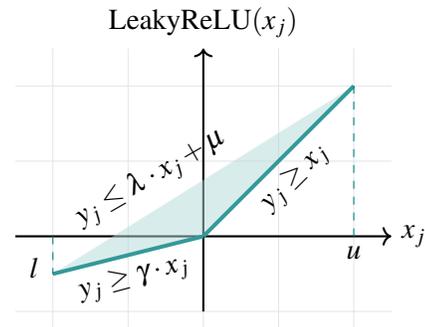
\begin{wrapfigure}[15]{r}{7cm}
    \centering
    \vspace*{-2.0ex}
    \begin{tikzpicture}
    \draw[gray!20,step=1] (-2.5,-1.25) grid (2.5,2.50);
    
    \draw[thick,->] (-2.5, 0) -- (2.5, 0) node[right] {$x_j$};
    \draw[thick,->] (0, -1.00) -- (0, 2.50) node[above] {$\text{LeakyReLU}(x_j)$};
    
    \fill[teal!30, opacity=0.5] (-2, -0.5) -- (2, 2) -- (0, 0);

    \draw (-2, -0.5) -- (0, 0) node[midway, below,rotate=15] {$y_j \geq \gamma \cdot x_j $};
    \draw[thick, teal!80,line width=1.5pt] (-2, -0.5) -- (0, 0);
    \draw (0, 0) -- (2, 2) node[midway, below,rotate=45] {$y_j \geq x_j$};
    \draw[thick, teal!80,line width=1.5pt] (0, 0) -- (2, 2);
    \draw (-1.55, -0.10) node[above,rotate=30,xshift=11.8mm] {$y_j \leq \lambda \cdot x_j + \mu $};
    
    \node[below] at (2, 0) {$u$};
    \node[below] at (-2.25, -0.15) {$l$};  

    \draw[teal, dashed] (2,0) -- (2,2);
    \draw[teal, dashed] (-2,0) -- (-2,-0.5);
\end{tikzpicture}
    \caption{Relaxation for the leaky ReLU function. The dark line shows the exact set and the light area the approximate set. In the figure, $\lambda = \frac{u - \gamma \cdot l}{u - l}$ and $ \mu = \frac{u \cdot l \cdot (\gamma - 1)}{u - l}$.}
    \label{fig:lrelu-relax}
\end{wrapfigure}   

\begin{figure}
    \centering
    \begin{subfigure}{.33\textwidth}
        \centering
        \begin{tikzpicture}[scale=0.8]
    \draw[gray!20,step=1] (-2.75, -0.8) grid (3.5, 3.5);
    
    \draw[thick,->] (-2.5, 0) -- (3, 0) node[right] {$x_j$};
    \draw[thick,->] (0, -0.5) -- (0, 3.75) node[above] {$\text{LeakyReLU}(x_j)$};
    
    \fill[teal!30, opacity=0.5] (-2.25, -0.5) -- (1.25, 3) -- (3,3) -- (0, 0) -- cycle;

    \draw (-2.25, -0.5) -- (0, 0) node[midway,below,rotate=15] {$y_j \geq x_j \cdot \gamma$};
    \draw[thick, teal!80,line width=1.5pt] (-2.25, -0.5) -- (0, 0);
    \draw (0, 0) -- (2, 2) node[midway,below,rotate=45] {$y_j \geq x_j$};
    \draw[thick, teal!80,line width=1.5pt] (0, 0) -- (3, 3);
    \draw (-1.60, 0.10) node[above,rotate=46,xshift=13.5mm] {$y_j \leq x_j + l \cdot\ (\gamma - 1) $};

    \node[below] at (-2.50, 0) {$l$};  
    \draw[teal, dashed] (-2.25,0) -- (-2.25,-0.50); 
\end{tikzpicture}
        \vspace*{-0.6cm}
        \caption{$x_j \in (l, \infty)$}
        \label{fig:lrelu-1}
    \end{subfigure}%
    \begin{subfigure}{.33\textwidth}
        \centering
        \begin{tikzpicture}[scale=0.8]
    \draw[gray!20,step=1] (-2.75, -0.8) grid (3.5, 3.5);
    
    \draw[thick,->] (-2.5, 0) -- (3, 0) node[right] {$x_j$};
    \draw[thick,->] (0, -0.5) -- (0, 3.75) node[above] {$\text{LeakyReLU}(x_j)$};
    
    \fill[teal!30, opacity=0.5] (-2.25, -0.5) -- (-2.25, 1.0555) -- (2,2) -- (0, 0) -- cycle;

    \draw (-2.25, -0.5) -- (0, 0) node[midway, below,rotate=12] {$y_j \geq x_j \cdot \gamma$};
    \draw[thick, teal!80,line width=1.5pt] (-2.25, -0.5) -- (0, 0);
    \draw (0, 0) -- (2, 2) node[midway, below,rotate=45] {$y_j \geq x_j$};
    \draw[thick, teal!80,line width=1.5pt] (0, 0) -- (3, 3);
    \draw (-1.85, 1.00) node[above,rotate=12,xshift=15mm] {$y_j \leq \gamma \cdot x_j + u \cdot (1 - \gamma) $};
    
    \node[below] at (2, 0) {$u$}; 
    \draw[teal, dashed] (2,0) -- (2,2); 
\end{tikzpicture}
        \vspace*{-0.6cm}
        \caption{$x_j \in (-\infty, u)$}
        \label{fig:lrelu-2}
    \end{subfigure}
    \begin{subfigure}{.33\textwidth}
        \centering
        \begin{tikzpicture}[scale=0.8]
    \draw[gray!20,step=1] (-2.75, -0.8) grid (3.5, 3.5);
    
    \draw[thick,->] (-2.5, 0) -- (3, 0) node[right] {$x_j$};
    \draw[thick,->] (0, -0.5) -- (0, 3.75) node[above] {$\text{LeakyReLU}(x_j)$};
    
    \fill[teal!30, opacity=0.5] (-2.25, -0.6) -- (-2.25, 3) -- (3,3) -- (0, 0) -- cycle;

    \draw (-2.25, -0.6) -- (0, 0) node[midway, below,rotate=10] {$y_j \geq x_j \cdot \gamma$};
    \draw[thick, teal!80,line width=1.5pt] (-2.25, -0.6) -- (0, 0);
    \draw (0, 0) -- (2, 2) node[midway, below,rotate=45] {$y_j \geq x_j$};
    \draw[thick, teal!80,line width=1.5pt] (0, 0) -- (3, 3);
\end{tikzpicture}
        \vspace*{-0.6cm}
        \caption{$x_j \in (-\infty, \infty)$}
        \label{fig:lrelu-3}
    \end{subfigure}
    \caption{Convex relaxations of the leaky ReLU function with three cases of an unbounded input set.}
    \label{fig:lrelu-unbounded}
\end{figure}
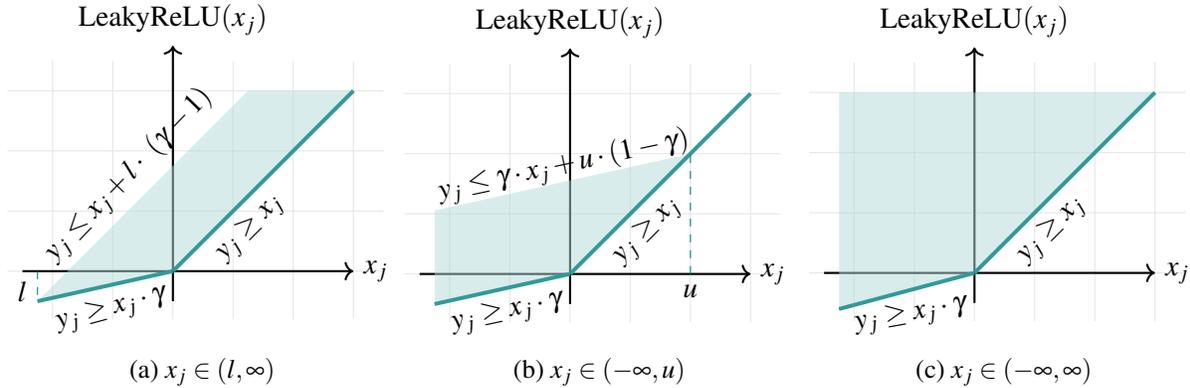

The over-approximate analysis of the leaky ReLU is also similar to the one for ReLU. For bounded inputs, correspondingly to the Planet relaxation \cite{DBLP:journals/pacmpl/SinghGPV19}, we also try to find an enclosing triangle, which is the tightest convex, linear relaxation that we can achieve for leaky ReLUs (see \autoref{fig:lrelu-relax}). 
The three constraints on the freshly introduced variable $\sPolyVars{m+1}$ are the following: (1) $\alpha_{m+1} \geq \gamma \cdot x_j$, (2) $\alpha_{m+1} \geq x_j$, and (3) $\alpha_{m+1} \leq \frac{u - \gamma \cdot l}{u - l}x_j + \frac{u \cdot l \cdot (\gamma - 1)}{u - l}$. At this point, the result of the $\actFunOA{j}{\text{L}}{\cdot}$ operation is a single star set with one more variable and three more constraints than the original input star. It is important to note: if the domain of variable $x_j$ is fully positive (i.e., $l \geq 0$) or fully negative (i.e., $u \leq 0$), then the resulting star is the same as described for the exact approach.
On the other hand, when there is an unbounded input set $\sSet$, three cases are distinguished: (1) $x_j \in (-\infty, u)$, (2) $x_j \in (l, \infty)$, and (3) $x_j \in (-\infty, \infty)$. The analysis for unbounded input is similar to the bounded case but the introduced constraints change, as visualized in \autoref{fig:lrelu-unbounded}. Note that these are the tightest linear, convex relaxations that can be achieved.

\subsection{Hard Tanh Layer}
\label{subsec:hardTanhlayer}
The hard hyperbolic tangent function, commonly known as the hard tanh function, is a linearized variant of the hyperbolic tangent activation function. In our work, we have generalized this function by introducing the parameters $V_{\text{min}}$ and $V_{\text{max}}$, which replace the original values of $-1$ and $1$, respectively \cite{collobert2004}. This modification allows us to flexibly adapt the function according to our specific needs and requirements.

\begin{definition}[Hard Hyperbolic Tangent] The \emph{hard hyperbolic tangent (HardTanh)} activation function with parameters $V_{\text{min}}\in\R$ and $V_{\text{max}}\in\R_{\geq V_{\text{min}}}$ is defined for each $x\in\R$ by
    \begin{equation}
        \label{eqn:modified-hardtanh}
            HardTanh(x) = \begin{cases}
                V_{\text{min}} & \textit{if } x < V_{\text{min}} \\
                x & \textit{if } V_{\text{min}} \leq x \leq V_{\text{max}}\\
                V_{\text{max}} & \textit{if } x > V_{\text{max}}\ .
            \end{cases}
        \end{equation}
\end{definition}


\subsubsection*{Exact Analysis}
\label{subsubsec:hardTanh-exact-analysis}
For the analysis of FNNs with the hard tanh activation function at neuron $n_j$, which we denote as $\actFun{j}{\text{H}}{\cdot}$, we split the result of the affine transformation $\sSet = \tuple{\sCenter{}, \sGeneratorM{}, \sPredicate}$ into three subsets: $\sSet_1=\tuple{\sCenter{}, \sGeneratorM{}, \sPredicate_1}$ is the intersection of $\sSet$ with the hyperplanes $V_{\text{min}} \leq x_j \leq V_{\text{max}}$, $\sSet_2=\tuple{\sCenter{}, \sGeneratorM{}, \sPredicate_2}$ with $x_j < V_{\text{min}}$ and $\sSet_3=\tuple{\sCenter{}, \sGeneratorM{}, \sPredicate_3}$ with $x_j > V_{\text{max}}$ (see Proposition \ref{prop:halfspace-intersect}).

According to \autoref{eqn:modified-hardtanh}, $\actFun{j}{\text{H}}{\cdot}$ leaves the elements of $\sSet_1$ unchanged since $x_j$ is in the range between $V_{\text{min}}$ and $V_{\text{max}}$.
For $\sSet_2$, all of its elements get the value $V_{\text{min}}$ in dimension $j$ since $x_j < V_{\text{min}}$, hence, we project the star onto $V_{\text{min}}$ in the dimension $j$. 
To achieve this result, we apply the mapping matrix $\matrixS{M} = [\vectorS{e}_1, \vectorS{e}_2, \hdots, \vectorS{e}_{j-1}, \vectorS{0}, \vectorS{e}_{j+1}, \hdots \vectorS{e}_n]$. Additionally, we set the $j^{\text{th}}$ dimension of the center to $V_{\text{min}}$ by adding the shifting vector $\matrixS{s}_{\text{min}} = [0, \hdots, V_{\text{min}}, \hdots, 0]^{\intercal}$ to the center.
For $\sSet_3$, we do the same by mapping the set with the mapping matrix, but instead, we set the center to $V_{\text{max}}$ by adding the shifting vector $\matrixS{s}_{\text{max}} = [0, \hdots, V_{\text{max}}, \hdots, 0]^{\intercal}$ to it. Thus, we project the star onto $V_{\text{max}}$ in the dimension $j$.
Accordingly, the $\actFun{j}{\text{H}}{\sSet}$ operation at neuron $j$ results in the union of three star sets: $\actFun{j}{\text{H}}{\cdot} = \tuple{\sCenter{}, \sGeneratorM{}, \sPredicate_1} \cup \tuple{\matrixS{M}\sCenter{}+\vectorS{s}_{\text{min}}, \matrixS{M}\sGeneratorM{}, \sPredicate_2} \cup \tuple{\matrixS{M}\sCenter{}+\vectorS{s}_{\text{max}}, \matrixS{M}\sGeneratorM{}, \sPredicate_3}$. 

Note that some of the intersections of the input star $\theta$ with the halfspaces $V_{\text{min}} \leq x_j \leq V_{\text{max}}$, $x_j < V_{\text{min}}$, and $x_j > V_{\text{max}}$ may be empty (see Proposition \ref{prop:emptiness}). In that case, we can spare the computation for the empty subsets, and continue the reachability analysis only with the non-empty resulting stars.

\subsubsection*{Over-approximate Analysis}
\label{subsubsec:hardTanh-overapprox-analysis}

In the over-approximate analysis, the $\actFunOA{j}{\text{H}}{\sSet}$ operation should yield a single star set. Thus we aim to find an enclosing triangle or trapezoid, which is the tightest convex, linear relaxation that we can achieve for hard tanh. For bounded inputs, we make a case distinction.
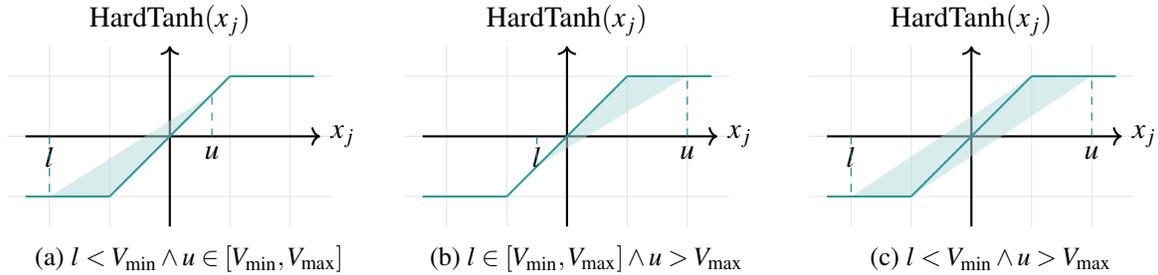
\begin{figure}[ht]
    \centering
    \begin{subfigure}{.33\textwidth}
        \centering
        \begin{tikzpicture}[scale=0.8]
    \draw[gray!20,step=1] (-2.7,-1.5) grid (2.7,1.5);

    \draw[thick,->] (-2.4, 0) -- (2.5, 0) node[right] {$x_j$};
    \draw[thick,->] (0, -1.5) -- (0, 1.5) node[above] {$\text{HardTanh}(x_j)$};

    \fill[teal!30, opacity=0.5] (-2, -1) -- (0.7, 0.7) -- (-1, -1);

    \draw[thick, teal!80] (-1, -1) -- (1, 1);
    \draw[thick, teal!80] (-2.4, -1) -- (-1, -1);
    \draw[thick, teal!80] (1, 1) -- (2.4, 1);

    \node[below] at (0.7, 0) {$u$};
    \node[below] at (-2, 0) {$l$};  

    \draw[teal, dashed] (0.7,0) -- (0.7,0.7);
    \draw[teal, dashed] (-2,0) -- (-2,-1);
\end{tikzpicture}
        \caption{$l < V_{\text{min}} \wedge u \in [V_{\text{min}}, V_{\text{max}}]$}
        \label{fig:htanh-1}
    \end{subfigure}%
    \begin{subfigure}{.33\textwidth}
        \centering
        \begin{tikzpicture}[scale=0.8]
    \draw[gray!20,step=1] (-2.7,-1.5) grid (2.7,1.5);
    
    \draw[thick,->] (-2.4, 0) -- (2.5, 0) node[right] {$x_j$};
    \draw[thick,->] (0, -1.5) -- (0, 1.5) node[above] {$\text{HardTanh}(x_j)$};
    
    \fill[teal!30, opacity=0.5] (-0.5, -0.5) -- (1, 1) -- (2, 1);

    \draw[thick, teal!80] (-1, -1) -- (1, 1);
    \draw[thick, teal!80] (-2.4, -1) -- (-1, -1);
    \draw[thick, teal!80] (1, 1) -- (2.4, 1);

    \node[below] at (2, 0) {$u$};
    \node[below] at (-0.5, 0) {$l$};  

    \draw[teal, dashed] (2,0) -- (2,1);
    \draw[teal, dashed] (-0.5,0) -- (-0.5,-0.5);
\end{tikzpicture}
        \caption{$l \in [V_{\text{min}}, V_{\text{max}}] \wedge u > V_{\text{max}}$}
        \label{fig:htanh-2}
    \end{subfigure}
    \begin{subfigure}{.33\textwidth}
        \centering
        \begin{tikzpicture}[scale=0.8]
    \draw[gray!20,step=1] (-2.7,-1.5) grid (2.7,1.5);
    
    \draw[thick,->] (-2.4, 0) -- (2.5, 0) node[right] {$x_j$};
    \draw[thick,->] (0, -1.5) -- (0, 1.5) node[above] {$\text{HardTanh}(x_j)$};
    
    \fill[teal!30, opacity=0.5] (-2, -1) -- (-1,-1) -- (2, 1) -- (1, 1);

    \draw[thick, teal!80] (-1, -1) -- (1, 1);
    \draw[thick, teal!80] (-2.4, -1) -- (-1, -1);
    \draw[thick, teal!80] (1, 1) -- (2.4, 1);

    \node[below] at (2, 0) {$u$};
    \node[below] at (-2, 0) {$l$};  

    \draw[teal, dashed] (2,0) -- (2,1);
    \draw[teal, dashed] (-2,0) -- (-2,-1);
\end{tikzpicture}
        \caption{$l < V_{\text{min}} \wedge u > V_{\text{max}}$}
        \label{fig:htanh-3}
    \end{subfigure}
    \caption{Relaxation for the hard tanh function. The dark line shows the exact set (non-convex) and the light area the approximate set (convex and linear).}
    \label{fig:htanh-overapprox}
\end{figure}
If the lower bound (in the bounding box of $\sSet$ in dimension $j$, see Proposition \ref{prop:bounding}) is less than $V_{\text{min}}$, and the upper bound is between $V_{\text{min}}$ and $V_{\text{max}}$, the three constraints on the newly introduced variable $\sPolyVars{m+1}$ are the following: (1) $\alpha_{m+1} \geq x_j$, (2) $\alpha_{m+1} \geq V_{\text{min}}$ and (3) $\alpha_{m+1} \leq \frac{u_j - V_{\text{min}}}{u_j - l_j}\cdot x_j - \frac{u_j \cdot (l_j - V_{\text{min}})}{u_j - l_j}$.
For the opposite case, we introduce the new variable $\sPolyVars{m+1}$ and three constraints: (1) $\alpha_{m+1} \leq V_{\text{max}}$, (2) $\alpha_{m+1} \leq x_j$ and (3) $\alpha_{m+1} \geq - \frac{l_j - V_{\text{max}}}{u_j - l_j} \cdot x_j - \frac{l_j \cdot (V_{\text{max}} - u_j)}{u_j - l_j} $.
When the star is over $V_{\text{min}}$ and $V_{\text{max}}$ (i.e., $l < V_{\text{min}} \land u > V_{\text{max}}$), we introduce the new variable $\sPolyVars{m+1}$ and four additional constraints: (1) $\alpha_{m+1} \geq V_{\text{min}}$, (2) $\alpha_{m+1} \leq V_{\text{max}}$, (3) $\alpha_{m+1} \leq \frac{V_{\text{max}} - V_{\text{min}}}{V_{\text{max}} - l_j} \cdot x_j - \frac{V_{\text{max}} \cdot (l_j - V_{\text{min}})}{V_{\text{max}} - l_j}$ and (4) $\alpha_{m+1} \geq \frac{V_{\text{min}} - V_{\text{max}}}{V_{\text{min}} - u_j} \cdot x_j - \frac{V_{\text{min}} \cdot (V_{\text{max}} - u_j)}{V_{\text{min}} - u_j}$.

It is important to highlight that when the domain of variable $x_j$ is between $V_{\text{min}}$ and $V_{\text{max}}$, less than $V_{\text{min}}$ (i.e., $u < V_{\text{min}}$) or greater than $V_{\text{max}}$  (i.e., $l > V_{\text{max}}$), the result is again a single star and is computed the same way as described in the exact approach.

Furthermore, when dealing with an unbounded input set $\sSet$ we distinguish three cases, as mentioned earlier. These cases are as follows: (1) $x_j \in (-\infty, u)$, (2) $x_j \in (l, \infty)$, and (3) $x_j \in (-\infty, \infty)$. The cases (1) and (2) are again divided into two sub-cases, hence we obtain five different cases, each one presented in \autoref{tab:unbounded-tanh}, coupled with the corresponding constraints and illustrations.

\begin{table}[ht!]
    \resizebox{1.0\textwidth}{!}{%
    \centering
    \begin{tabularx}{\textwidth}{|c|>{\centering\arraybackslash}X|c|}
        \hline
        Domain of $x_j$ & Introduced constraints & Graphical illustration \\
        \hline
        \begin{minipage}{5cm}
            \centering
            \vspace*{-2ex}
            $l = -\infty \wedge u \in [V_{\text{min}}, V_{\text{max}}]$
        \end{minipage} & 
        \begin{minipage}[c]{\linewidth}
            \vspace*{-2ex}
            $\alpha_{m+1} \geq V_{\text{min}}$ \\
            $\alpha_{m+1} \geq x_j$ \\
            $\alpha_{m+1} \leq u_j$ 
        \end{minipage} & 
        \begin{tikzpicture}[baseline=0,scale=0.5]
            \draw[gray!20,step=1] (-2.7,-1.5) grid (2.7,1.5);
            
            \draw[thick,->] (-2.4, 0) -- (2.4, 0) node[right] {\scalebox{0.7}{$x_j$}};
            \draw[thick,->] (0, -1.5) -- (0, 1.5) node[above] {\scalebox{0.7}{$y_j$}};
            
            \fill[teal!30, opacity=0.5] (-2.4, -1) -- (-2.4,0.7) -- (0.7, 0.7) -- (-1, -1) -- cycle;

            \draw[thick, teal!80] (-1, -1) -- (1, 1);
            \draw[thick, teal!80] (-2.4, -1) -- (-1, -1);
            \draw[thick, teal!80] (1, 1) -- (2.4, 1);

            \foreach \x in {-2,-1,1,2}
                \draw (\x,-0.1) -- (\x,0.1);
            \foreach \y in {-1,1}
                \draw (-0.1,\y) -- (0.1,\y);

            \node[below] at (0.7, 0) {\scalebox{0.7}{$u$}};

            \draw[teal, dashed] (0.7,0) -- (0.7,0.7);
            \addvmargin{1mm}
        \end{tikzpicture} \\
        \hline
            \begin{minipage}{5cm}
                \centering
                \vspace*{-2ex}
                $l = -\infty \wedge u > V_{\text{max}}$
            \end{minipage} & 
            \begin{minipage}[Ht]{\linewidth}
                \vspace*{-2ex}
                $\alpha_{m+1} \geq V_{\text{min}}$ \\
                $\alpha_{m+1} \leq V_{\text{max}}$ \\
                $\alpha_{m+1} \geq \frac{V_{\text{min}} - V_{\text{max}}}{V_{\text{min}} - u_j} \cdot x_j - \frac{V_{\text{min}} \cdot (V_{\text{max}} - u_j)}{V_{\text{min}} - u_j}$
            \end{minipage} & 
            \begin{tikzpicture}[baseline=0,scale=0.5]
                \draw[gray!20,step=1] (-2.7,-1.5) grid (2.7,1.5);
                
            \draw[thick,->] (-2.4, 0) -- (2.4, 0) node[right] {\scalebox{0.7}{$x_j$}};
            \draw[thick,->] (0, -1.5) -- (0, 1.5) node[above] {\scalebox{0.7}{$y_j$}};
                
                \fill[teal!30, opacity=0.5] (-2.4, -1) -- (-2.4,1) -- (2, 1) -- (-1, -1) -- cycle;

                \draw[thick, teal!80] (-1, -1) -- (1, 1);
                \draw[thick, teal!80] (-2.4, -1) -- (-1, -1);
                \draw[thick, teal!80] (1, 1) -- (2.4, 1);

                \foreach \x in {-2,-1,1,2}
                    \draw (\x,-0.1) -- (\x,0.1);
                \foreach \y in {-1,1}
                    \draw (-0.1,\y) -- (0.1,\y);

                \node[below] at (2, 0) {\scalebox{0.7}{$u$}};

                \draw[teal, dashed] (2,0) -- (2,1);

                \addvmargin{1mm}
            \end{tikzpicture} \\
        \hline
            \begin{minipage}{5cm}
                \centering
                \vspace*{-2ex}
                $l \in [V_{\text{min}}, V_{\text{max}}] \wedge u = \infty$
            \end{minipage} & 
            \begin{minipage}[Ht]{\linewidth}
                \vspace*{-2ex}
                $\alpha_{m+1} \leq V_{\text{max}}$ \\
                $\alpha_{m+1} \leq x_j$ \\
                $\alpha_{m+1} \geq l_j$
            \end{minipage} & 
            \begin{tikzpicture}[baseline=0,scale=0.5]
                \draw[gray!20,step=1] (-2.7,-1.5) grid (2.7,1.5);
                
                \draw[thick,->] (-2.4, 0) -- (2.4, 0) node[right] {\scalebox{0.7}{$x_j$}};
                \draw[thick,->] (0, -1.5) -- (0, 1.5) node[above] {\scalebox{0.7}{$y_j$}};
                
                \fill[teal!30, opacity=0.5] (-0.7, -0.7) -- (1, 1) -- (2.4,1) -- (2.4, -0.7) -- cycle;

                \draw[thick, teal!80] (-1, -1) -- (1, 1);
                \draw[thick, teal!80] (-2.4, -1) -- (-1, -1);
                \draw[thick, teal!80] (1, 1) -- (2.4, 1);

                \foreach \x in {-2,-1,1,2}
                    \draw (\x,-0.1) -- (\x,0.1);
                \foreach \y in {-1,1}
                    \draw (-0.1,\y) -- (0.1,\y); 

                \draw[teal, dashed] (-0.7,0) -- (-0.7,-0.7);

                \node[below] at (-0.7, -0.7) {\scalebox{0.7}{$l$}}; 
                \addvmargin{1mm}
            \end{tikzpicture} \\
        \hline 
            \begin{minipage}{5cm}
                \centering
                \vspace*{-2ex}
                $l < V_{\text{min}} \wedge u = \infty$ 
            \end{minipage} &
            \begin{minipage}[Ht]{\linewidth}
                \vspace*{-2ex}
                $\alpha_{m+1} \leq V_{\text{max}}$ \\
                $\alpha_{m+1} \geq V_{\text{min}}$ \\
                $\alpha_{m+1} \leq \frac{V_{\text{max}} - V_{\text{min}}}{V_{\text{max}} - l_j}\cdot x_j - \frac{V_{\text{max}} \cdot (l_j - V_{\text{min}})}{V_{\text{max}} - l_j}$
            \end{minipage} & 
            \begin{tikzpicture}[baseline=0,scale=0.5]
                \draw[gray!20,step=1] (-2.7,-1.5) grid (2.7,1.5);
                
                \draw[thick,->] (-2.4, 0) -- (2.4, 0) node[right] {\scalebox{0.7}{$x_j$}};
                \draw[thick,->] (0, -1.5) -- (0, 1.5) node[above] {\scalebox{0.7}{$y_j$}};
                
                \fill[teal!30, opacity=0.5] (-2, -1) -- (2.4,-1) -- (2.4, 1) -- (1, 1);

                \draw[thick, teal!80] (-1, -1) -- (1, 1);
                \draw[thick, teal!80] (-2.4, -1) -- (-1, -1);
                \draw[thick, teal!80] (1, 1) -- (2.4, 1);

                \foreach \x in {-2,-1,1,2}
                    \draw (\x,-0.1) -- (\x,0.1);
                \foreach \y in {-1,1}
                    \draw (-0.1,\y) -- (0.1,\y); 

                \draw[teal, dashed] (-2,0) -- (-2,-1);

                \node[below] at (-2, 0) {\scalebox{0.7}{$l$}}; 
                \addvmargin{1mm}
            \end{tikzpicture} \\
        \hline
            \begin{minipage}{5cm}
                \centering
                \vspace*{-2ex}
                $l = -\infty \wedge u = \infty$
            \end{minipage} & 
            \begin{minipage}[Ht]{\linewidth}
                \vspace*{-2ex}
                $\alpha_{m+1} \geq V_{\text{min}}$ \\
                $\alpha_{m+1} \leq V_{\text{max}}$
            \end{minipage} & 
            \begin{tikzpicture}[baseline=0,scale=0.5]
                \draw[gray!20,step=1] (-2.7,-1.5) grid (2.7,1.5);
                
                \draw[thick,->] (-2.4, 0) -- (2.4, 0) node[right] {\scalebox{0.7}{$x_j$}};
                \draw[thick,->] (0, -1.5) -- (0, 1.5) node[above] {\scalebox{0.7}{$y_j$}};
                
                \fill[teal!30, opacity=0.5] (-2.4, -1) -- (-2.4,1) -- (2.4, 1) -- (2.4, -1) -- cycle;

                \draw[thick, teal!80] (-1, -1) -- (1, 1);
                \draw[thick, teal!80] (-2.4, -1) -- (-1, -1);
                \draw[thick, teal!80] (1, 1) -- (2.4, 1);

                \foreach \x in {-2,-1,1,2}
                    \draw (\x,-0.1) -- (\x,0.1);
                \foreach \y in {-1,1}
                    \draw (-0.1,\y) -- (0.1,\y);

                \addvmargin{1mm}
            \end{tikzpicture} \\
        \hline
    \end{tabularx}
    \caption{Approximation rules for the hard tanh function, when the input is unbounded. We distinguish five cases in total, for each we show the case itself, the introduced constraints and a graphical illustration.}
    \label{tab:unbounded-tanh}
}

\end{table}

\subsection{Hard Sigmoid Layer}
\label{subsec:hardSigmoid-layer}
The hard sigmoid activation function is a linearized variant of the sigmoid function. Since the hard sigmoid function has different variants in use \cite{tensorflowhsigmoid,anthadupula2021review,pytorch2021}, we generalize it by adding parameters.

\begin{definition}[Hard Sigmoid Function]
    The \emph{hard sigmoid (HardSigmoid)} function with parameters $V_{\text{min}}\in\R$ and $V_{\text{max}}\in\R_{\geq V_{\text{min}}}$ is defined for each $x\in\R$ by
    \begin{equation}\label{eqn:modified-hardsigmoid}
        HardSigmoid(x)=
        \begin{cases}
            0 & \textit{if } x \leq V_{\text{min}}\\
            \frac{1}{V_{\text{max}} - V_{\text{min}}}\cdot x + \frac{V_{\text{min}}}{V_{\text{min}} - V_{\text{max}}} & \textit{if }  V_{\text{min}} < x < V_{\text{max}}\\
            1 & \textit{if } x \geq V_{\text{max}}  \ .
            \end{cases}
    \end{equation}
\end{definition}

\subsubsection*{Exact Analysis}
\label{subsubsec:hardSigmoid-exact-analysis}
The analysis of the hard sigmoid works similarly to the one of the hard tanh function. The difference is that instead of the star remaining the same in the range between $V_{\text{min}}$ and $V_{\text{max}}$, we scale the star according to \autoref{eqn:modified-hardsigmoid}.
To compute $\actFun{j}{\text{S}}{\sSet}$, the star $\sSet = \tuple{\sCenter{}, \sGeneratorM{}, \sPredicate}$ 
is partitioned into three subsets $\sSet_1$, $\sSet_2$ and $\sSet_3$, covering the partitions with $V_{\text{min}} < x_j < V_{\text{max}}$, $x_j \leq V_{\text{min}}$ respectively $x_j \geq V_{\text{max}}$.
We scale $\sSet_1$ by applying the scaling matrix $\matrixS{M}_{sc} = [\vectorS{e}_1, \vectorS{e}_2, \hdots, \frac{1}{V_{\text{max}} - V_{\text{min}}} \vectorS{e}_{j}, \hdots \vectorS{e}_{n-1}, \vectorS{e}_n]$ and shift the center with the translation vector $\matrixS{s}_{sc} = [0, \hdots, \frac{V_{\text{min}}}{V_{\text{min}} - V_{\text{max}}}, \hdots, 0]^{\intercal}$.
Furthermore, the elements of $\sSet_2$ are set to zero in dimension $j$ by applying the mapping matrix $\matrixS{M} = [\vectorS{e}_1, \vectorS{e}_2, \hdots, \vectorS{e}_{j-1}, \vectorS{0}, \vectorS{e}_{j+1}, \hdots \vectorS{e}_n]$.
Finally, the elements of $\sSet_3$ are set to one by using the same projection $\matrixS{M}$, plus setting the center to one by the shifting vector $\matrixS{s}_{one} = [0, \hdots, 1, \hdots, 0]^{\intercal}$.
Consequently, the result is the union of three stars: $\actFun{j}{\text{S}}{\sSet} = \tuple{\matrixS{M}_{sc}\sCenter{} + \vectorS{s}_{sc},\, \matrixS{M}_{sc}\sGeneratorM{},\, \sPredicate_1} \cup \tuple{\matrixS{M}\sCenter{},\, \matrixS{M}\sGeneratorM{},\, \sPredicate_2} \cup \tuple{\matrixS{M}\sCenter{}+\vectorS{s}_{one},\, \matrixS{M}\sGeneratorM{},\, \sPredicate_3}$. 

Again, when intersecting the star $\sSet$ with $V_{\text{min}} < x_j < V_{\text{max}}$, $x_j \leq V_{\text{min}}$ respectively $x_j \geq V_{\text{max}}$, certain resulting subsets may become empty (see \ref{prop:emptiness}) and thus their further processing can be omitted.

\subsubsection*{Over-approximate Analysis}
\label{subsubsec:hardSigmoid-overapprox-analysis}
Using the over-approximate analysis of hard sigmoid, we consider cases where a convex triangle or trapezoid is applicable based on the input.
The $\actFunOA{j}{\text{S}}{\sSet}$ operation introduces a new variable $\sPolyVars{m+1}$ regardless of which case occurs.
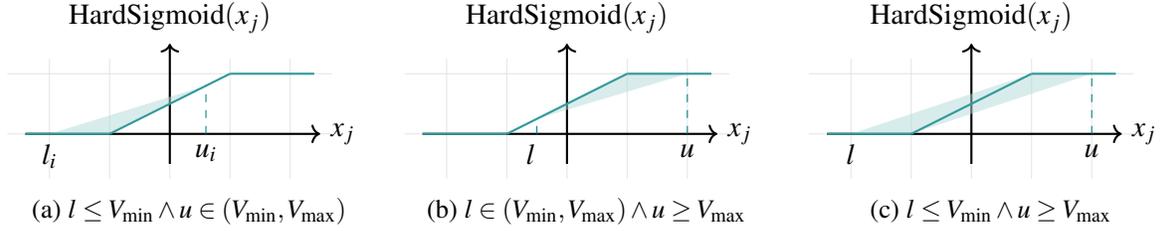
\begin{figure}[t]
    \centering
    \begin{subfigure}{.33\textwidth}
        \centering
        \begin{tikzpicture}[scale=0.8]
    \draw[gray!20,step=1] (-2.7,-0.75) grid (2.7,1.25);
    
    \draw[thick,->] (-2.4, 0) -- (2.5, 0) node[right] {$x_j$};
    \draw[thick,->] (0, -0.5) -- (0, 1.5) node[above] {$\text{HardSigmoid}(x_j)$};
        
        \fill[teal!30, opacity=0.5] (-2, 0) -- (0.6, 0.8) -- (-1, 0);
        
        \draw[thick, teal!80] (-1, 0) -- (1, 1);
        \draw[thick, teal!80] (-2.4, 0) -- (-1, 0);
        \draw[thick, teal!80] (1, 1) -- (2.4, 1);

        \node[below] at (0.6, 0) {$u_i$};
        \node[below] at (-2, 0) {$l_i$};  

        \draw[teal, dashed] (0.6,0) -- (0.6,0.8);
\end{tikzpicture}
        \caption{$l \leq V_{\text{min}} \wedge u \in (V_{\text{min}}, V_{\text{max}})$}
        \label{fig:hsigmoid-1}
    \end{subfigure}%
    \begin{subfigure}{.33\textwidth}
        \centering
        \begin{tikzpicture}[scale=0.8]
    \draw[gray!20,step=1] (-2.7,-0.75) grid (2.7,1.25);
    
    \draw[thick,->] (-2.4, 0) -- (2.5, 0) node[right] {$x_j$};
    \draw[thick,->] (0, -0.5) -- (0, 1.5) node[above] {$\text{HardSigmoid}(x_j)$};
        
    \fill[teal!30, opacity=0.5] (-0.5, 0.25) -- (1, 1) -- (2, 1);

    \draw[thick, teal!80] (-1, 0) -- (1, 1);
    \draw[thick, teal!80] (-2.4, 0) -- (-1, 0);
    \draw[thick, teal!80] (1, 1) -- (2.4, 1);

    \node[below] at (2, 0) {$u$};
    \node[below] at (-0.6, 0) {$l$};  

    \draw[teal, dashed] (2,0) -- (2,1);
    \draw[teal, dashed] (-0.5,0) -- (-0.5,0.25);
\end{tikzpicture}
        \caption{$l \in (V_{\text{min}}, V_{\text{max}}) \wedge u \geq V_{\text{max}}$}
        \label{fig:hsigmoid-2}
    \end{subfigure}
    \begin{subfigure}{.33\textwidth}
        \centering
        \begin{tikzpicture}[scale=0.8]
    \draw[gray!20,step=1] (-2.7,-0.75) grid (2.7,1.25);
    
    \draw[thick,->] (-2.4, 0) -- (2.5, 0) node[right] {$x_j$};
    \draw[thick,->] (0, -0.5) -- (0, 1.5) node[above] {$\text{HardSigmoid}(x_j)$};
        
    \fill[teal!30, opacity=0.5] (-2, 0) -- (-1,0) -- (2, 1) -- (1, 1);

    \draw[thick, teal!80] (-1, 0) -- (1, 1);
    \draw[thick, teal!80] (-2.4, 0) -- (-1, 0);
    \draw[thick, teal!80] (1, 1) -- (2.4, 1);

    \node[below] at (2, 0) {$u$};
    \node[below] at (-2, 0) {$l$};  

    \draw[teal, dashed] (2,0) -- (2,1);
\end{tikzpicture}
        \caption{$l \leq V_{\text{min}} \wedge u \geq V_{\text{max}}$}
        \label{fig:hsigmoid-3}
    \end{subfigure}
    \caption{Relaxation for the hard tanh function. The dark line shows the exact set (non-convex) and the light area the approximate set (convex and linear).}
    \label{fig:hsigmoid-overapprox}
\end{figure} 

If the lower bound is less than $V_{\text{min}}$ and the upper bound is between $V_{\text{min}}$ and $V_{\text{max}}$, then three new constraints are introduced: (1) $\alpha_{m+1} \geq 0$, (2) $\alpha_{m+1} \geq \frac{1}{V_{\text{max}} - V_{\text{min}}}\cdot x_j + \frac{V_{\text{min}}}{V_{\text{max}} - V_{\text{min}}}$, and (3) $\alpha_{m+1} \leq \frac{u \cdot (x_j - l)}{u - l}$.
In the dual scenario when the lower bound is between $V_{\text{min}}$ and $V_{\text{max}}$ while the upper bound exceeds $V_{\text{max}}$, we encounter the constraints: (1) $\alpha_{m+1} \leq 1$, (2) $\alpha_{m+1} \leq \frac{1}{V_{\text{max}} - V_{\text{min}}}\cdot x_i - \frac{V_{\text{min}}}{V_{\text{max}} - V_{\text{min}}}$, and (3) $\alpha_{m+1} \geq \frac{l -1}{l - u}\cdot x_j + \frac{l \cdot (1 - u)}{l - u}$.
Lastly, when in dimension $j$ the star is between $V_{\text{min}}$ and $V_{\text{max}}$, then we introduce four constraints: (1) $\alpha_{m+1} \leq 1$, (2) $\alpha_{m+1} \geq 0$, (3) $\alpha_{m+1} \leq \frac{1}{V_{\text{max}} - l}\cdot x_j - \frac{l}{V_{\text{max}} - l}$, and (4) $\alpha_{m+1} \geq \frac{1}{V_{\text{min}} - u}\cdot x_j - \frac{V_{\text{min}}}{u - V_{\text{min}}}$.
It is important to highlight that when the domain of variable $x_j$ is between $V_{\text{min}}$ and $V_{\text{max}}$, less than $V_{\text{min}}$ (i.e., $u \leq V_{\text{min}}$) or greater than $V_{\text{max}}$  (i.e., $l \geq V_{\text{max}}$), the resulting stars remain the same as described in the exact approach.

Furthermore, when dealing with an unbounded input set $\sSet$ we distinguish three cases, as mentioned earlier. These cases are as follows: (1) $x_j \in (-\infty, u)$, (2) $x_j \in (l, \infty)$, and (3) $x_j \in (-\infty, \infty)$. The cases (1) and (2) are again divided into two sub-cases, hence we obtain five different cases, each one presented in \autoref{tab:unbounded-sigmoid}, coupled with the corresponding constraints and illustrations.

\begin{table}[ht!]
    \resizebox{1.0\textwidth}{!}{%
    \centering
    \begin{tabularx}{\textwidth}{|c|>{\centering\arraybackslash}X|c|}
        \hline
        Domain of $x_j$ & Introduced constraints & Graphical illustration \\
        \hline
        \begin{minipage}{5cm}
            \centering
            \vspace*{-3ex}
            $l = -\infty \wedge u \in [V_{\text{min}}, V_{\text{max}}]$
        \end{minipage} & 
        \begin{minipage}[c]{\linewidth}
            \vspace*{-3ex}
            $\alpha_{m+1} \geq V_{\text{min}}$ \\
            $\alpha_{m+1} \geq x_j$ \\
            $\alpha_{m+1} \leq u_j$ 
        \end{minipage} & 
        
        \begin{tikzpicture}[baseline=0,scale=0.5]
            \draw[gray!20,step=1] (-2.7,-0.75) grid (2.7,1.75);
                
            \draw[thick,->] (-2.4, 0) -- (2.4, 0) node[right] {\scalebox{0.7}{$x_j$}};
            \draw[thick,->] (0, -0.5) -- (0, 1.5) node[above] {\scalebox{0.7}{$y_j$}};
            
            \fill[teal!30, opacity=0.5] (-2.4, 0) -- (-2.4,0.85) -- (0.7 , 0.85) -- (-1, 0) -- cycle;
    
            \draw[thick, teal!80] (-1, 0) -- (1, 1);
            \draw[thick, teal!80] (-2.4, 0) -- (-1, 0);
            \draw[thick, teal!80] (1, 1) -- (2.4, 1);

            \foreach \x in {-2,-1,1,2}
                \draw (\x,-0.1) -- (\x,0.1);
            \foreach \y in {1}
                \draw (-0.1,\y) -- (0.1,\y);

            \node[below] at (0.7, 0) {\scalebox{0.7}{$u$}};

            \draw[teal, dashed] (0.7,0) -- (0.7,0.85);

            \addvmargin{1mm}
        \end{tikzpicture} \\
        \hline
            \begin{minipage}{5cm}
                \centering
                \vspace*{-3ex}
                $l = -\infty \wedge u > V_{\text{max}}$
            \end{minipage} & 
            \begin{minipage}[c]{\linewidth}
                \vspace*{-3ex}
                $\alpha_{m+1} \geq V_{\text{min}}$ \\
                $\alpha_{m+1} \leq V_{\text{max}}$ \\
                $\alpha_{m+1} \geq \frac{V_{\text{min}} - V_{\text{max}}}{V_{\text{min}} - u_j} \cdot x_j - \frac{V_{\text{min}} \cdot (V_{\text{max}} - u_j)}{V_{\text{min}} - u_j}$
            \end{minipage} & 
            \begin{tikzpicture}[baseline=0,scale=0.5]
                \draw[gray!20,step=1] (-2.7,-0.75) grid (2.7,1.75);
                
                \draw[thick,->] (-2.4, 0) -- (2.4, 0) node[right] {\scalebox{0.7}{$x_j$}};
                \draw[thick,->] (0, -0.5) -- (0, 1.5) node[above] {\scalebox{0.7}{$y_j$}};
                 
                \fill[teal!30, opacity=0.5] (-2.4, 0) -- (-2.4,1) -- (2, 1) -- (-1, 0) -- cycle;
        
                \draw[thick, teal!80] (-1, 0) -- (1, 1);
                \draw[thick, teal!80] (-2.4, 0) -- (-1, 0);
                \draw[thick, teal!80] (1, 1) -- (2.4, 1);

                \foreach \x in {-2,-1,1,2}
                    \draw (\x,-0.1) -- (\x,0.1);
                \foreach \y in {1}
                    \draw (-0.1,\y) -- (0.1,\y);

                \node[below] at (2, 0) {\scalebox{0.7}{$u$}};

                \draw[teal, dashed] (2,0) -- (2,1);

                \addvmargin{1mm}
            \end{tikzpicture} \\
        \hline
            \begin{minipage}{5cm}
                \centering
                \vspace*{-3ex}
                $l \in [V_{\text{min}}, V_{\text{max}}] \wedge u = \infty$ 
            \end{minipage} & 
            \begin{minipage}[c]{\linewidth}
                \vspace*{-3ex}
                $\alpha_{m+1} \leq V_{\text{max}}$ \\
                $\alpha_{m+1} \leq x_j$ \\
                $\alpha_{m+1} \geq l_j$
            \end{minipage} & 
            \begin{tikzpicture}[baseline=0,scale=0.5]
                \draw[gray!20,step=1] (-2.7,-0.75) grid (2.7,1.75);
                
                \draw[thick,->] (-2.4, 0) -- (2.4, 0) node[right] {\scalebox{0.7}{$x_j$}};
                \draw[thick,->] (0, -0.5) -- (0, 1.5) node[above] {\scalebox{0.7}{$y_j$}};
                
            \fill[teal!30, opacity=0.5] (-0.5, 0.25) -- (1, 1) -- (2.4,1) -- (2.4, 0.25) -- cycle;
    
            \draw[thick, teal!80] (-1, 0) -- (1, 1);
            \draw[thick, teal!80] (-2.4, 0) -- (-1, 0);
            \draw[thick, teal!80] (1, 1) -- (2.4, 1);

                \foreach \x in {-2,-1,1,2}
                    \draw (\x,-0.1) -- (\x,0.1);
                \foreach \y in {1}
                    \draw (-0.1,\y) -- (0.1,\y);

                \node[below] at (-0.5, 0) {\scalebox{0.7}{$l$}};  

                \draw[teal, dashed] (-0.5,0) -- (-0.5,0.25);
                \addvmargin{1mm}
            \end{tikzpicture} \\
        \hline 
            \begin{minipage}{5cm}
                \centering
                \vspace*{-3ex}
                $l < V_{\text{min}} \wedge u = \infty$
            \end{minipage} & 
            \begin{minipage}[c]{\linewidth}
                \vspace*{-3ex}
                $\alpha_{m+1} \leq V_{\text{max}}$ \\
                $\alpha_{m+1} \geq V_{\text{min}}$ \\
                $\alpha_{m+1} \leq \frac{V_{\text{max}} - V_{\text{min}}}{V_{\text{max}} - l_j}\cdot x_j - \frac{V_{\text{max}} \cdot (l_j - V_{\text{min}})}{V_{\text{max}} - l_j}$
            \end{minipage} & 
            \begin{tikzpicture}[baseline=0,scale=0.5]
                \draw[gray!20,step=1] (-2.7,-0.75) grid (2.7,1.75);
                
                \draw[thick,->] (-2.4, 0) -- (2.4, 0) node[right] {\scalebox{0.7}{$x_j$}};
                \draw[thick,->] (0, -0.5) -- (0, 1.5) node[above] {\scalebox{0.7}{$y_j$}};
                
                \fill[teal!30, opacity=0.5] (-2, 0) -- (1, 1) -- (2.4,1) -- (2.4, 0) -- cycle;
        
                \draw[thick, teal!80] (-1, 0) -- (1, 1);
                \draw[thick, teal!80] (-2.4, 0) -- (-1, 0);
                \draw[thick, teal!80] (1, 1) -- (2.4, 1);

                \foreach \x in {-2,-1,1,2}
                    \draw (\x,-0.1) -- (\x,0.1);
                \foreach \y in {1}
                    \draw (-0.1,\y) -- (0.1,\y);
        
                \node[below] at (-2, 0) {\scalebox{0.7}{$l$}}; 

                \addvmargin{1mm}
            \end{tikzpicture} \\
        \hline
            \begin{minipage}{5cm}
                \centering
                \vspace*{-3ex}
                $l = -\infty \wedge u = \infty$
            \end{minipage} & 
            \begin{minipage}[c]{\linewidth}
                \vspace*{-3ex}
                $\alpha_{m+1} \geq V_{\text{min}}$ \\
                $\alpha_{m+1} \leq V_{\text{max}}$
            \end{minipage} & 
            \begin{tikzpicture}[baseline=0,scale=0.5]
                \draw[gray!20,step=1] (-2.7,-0.75) grid (2.7,1.75);
                
                \draw[thick,->] (-2.4, 0) -- (2.4, 0) node[right] {\scalebox{0.7}{$x_j$}};
                \draw[thick,->] (0, -0.5) -- (0, 1.5) node[above] {\scalebox{0.7}{$y_j$}};
                
                \fill[teal!30, opacity=0.5] (-2.4, 0) -- (-2.4,1) -- (2.4, 1) -- (2.4, -0) -- cycle;

                \draw[thick, teal!80] (-1, 0) -- (1, 1);
                \draw[thick, teal!80] (-2.4, 0) -- (-1, 0);
                \draw[thick, teal!80] (1, 1) -- (2.4, 1);

                \foreach \x in {-2,-1,1,2}
                    \draw (\x,-0.1) -- (\x,0.1);
                \foreach \y in {1}
                    \draw (-0.1,\y) -- (0.1,\y);

                \addvmargin{1mm}
            \end{tikzpicture} \\
        \hline
    \end{tabularx}
    \caption{Approximation rules for hard sigmoid, when the input is unbounded. In total, we distinguish five cases, for each we show the case itself, the introduced constraints and a graphical illustration.}
    \label{tab:unbounded-sigmoid}
}
\end{table}

\subsection{Unit Step Function Layer}
\label{subsec:unitStep-layer}

The unit step activation function (also called the heaviside function) is widely used in neural networks. In this work, we generalize the unit step function, by introducing three parameters with commonly used values $val = 0$, $R_{\text{min}} = 0$, and $R_{\text{max}}=1$. 

\begin{definition}[Unit Step \cite{stepfunction}] The \emph{unit step} function with \emph{separator} $val\in \R$, \emph{lower limit} $R_{\text{min}}\in\R$ and \emph{upper limit} $R_{\text{max}}\in\R_{\geq R_{\text{min}}}$ is defined for each $x\in \R$ by
    \begin{equation}\label{mystepfunction}
        \mathit{UnitStep}(x)=
        \begin{cases}
            R_{\text{min}} & \textit{if }  x < val\\
            R_{\text{max}} & \textit{if } x \geq val\ .
        \end{cases}
    \end{equation}
\end{definition}



\subsubsection*{Exact Analysis}
\label{subsubsec:unitStep-exact-analysis}

The result $\actFun{j}{\text{U}}{\sSet}$ of applying unit step on a star $\sSet = \tuple{\sCenter{}, \sGeneratorM{}, \sPredicate}$ in dimension $j$ is obtained as follows. First, $\sSet$ is decomposed into two parts $\sSet_1$ and $\sSet_2$ that result from the intersection of $\sSet$ with $x_j < \mathit{val}$ resp. $x_j \geq \mathit{val}$. Then, the values in the $j$th dimension are set to $R_{\text{min}}$ and $R_{\text{max}}$, respectively in the stars $\sSet_1$ and $\sSet_2$. We achieve this by using the projection matrix $\matrixS{M} = [\vectorS{e}_1, \vectorS{e}_2, \hdots, \vectorS{e}_{j-1}, \vectorS{0}, \vectorS{e}_{j+1}, \hdots \vectorS{e}_n]$ and translation vectors $\vectorS{s}_{\text{min}} = [0, \hdots, R_{\text{min}}, \hdots, 0]^{\intercal}$ and $\vectorS{s}_{\text{max}} = [0, \hdots, R_{\text{min}}, \hdots, 0]^{\intercal}$. The resulting stars are $\tuple{\matrixS{M} \sCenter{}+\vectorS{s}_{\text{min}},\, \matrixS{M} \sGeneratorM{},\, \sPredicate_1}$ and $\tuple{ \matrixS{M} \sCenter{}+\vectorS{s}_{\text{max}},\, \matrixS{M} \sGeneratorM{},\, \sPredicate_2}$. Note that if the domain $(l, u)$ of $x_j$ does not contain the value $val$, then the case splitting is not necessary and only one of the stars is the final result, correspondingly to the non-empty intersection with one of the halfspaces.

\subsubsection*{Over-approximate Analysis}
\label{subsubsec:unitStep-overapprox-analysis}

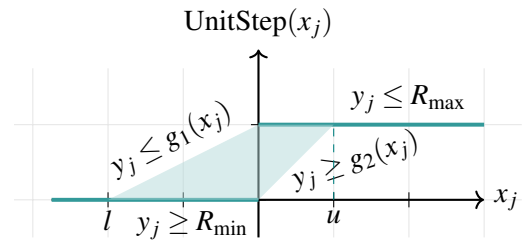
\begin{wrapfigure}[13]{r}{7cm}
    \centering
    \vspace*{-1.0ex}
    \begin{tikzpicture}
    \draw[gray!20,step=1] (-3.25,-0.5) grid (3.5,1.75);
    
    \draw[thick,->] (-2.75, 0) -- (3, 0) node[right] {$x_j$};
    \draw[thick,->] (0, -0.5) -- (0, 2) node[above] {$\text{UnitStep}(x_j)$};
    
    \foreach \x in {-2,-1,1,2}
        \draw (\x,-0.1) -- (\x,0.1);
    \foreach \y in {0,1}
        \draw (-0.1,\y) -- (0.1,\y);
    
    \draw[line width=1.5pt, teal!80] (-2.75, 0) -- (0, 0) node [midway, below, xshift=0.5cm, color=black] {$y_j \geq R_{\text{min}}$};
    \draw[line width=1.5pt, teal!80] (0, 1) -- (3, 1) node [midway, above, xshift=0.5cm, color=black] {$y_j \leq R_{\text{max}}$};
    \draw (-1, 0.5) node[above,rotate=27] {$y_j \leq  g_1(x_j)$};
    \draw (+1.4, +0.20) node[above,rotate=18] {$y_j \geq  g_2(x_j)$};

    \fill[teal!30, opacity=0.5] (-2, 0) -- (0, 1) -- (1, 1) -- (0, 0);
    
    \draw[teal, dashed] (1,0) -- (1,1);

    \node[below] at (1, 0) {$u$};
    \node[below] at (-2, 0) {$l$};
\end{tikzpicture}
    \caption{Relaxation for the unit step function. The dark line shows the exact set and the light area the approximate set. The constraints $y_j \leq g_1(x_j)$ and $y_j \geq g_2(x_j)$ correspond to relaxations (3) and (4).}
    \label{fig:unitstep-relax}
\end{wrapfigure}
The over-approximate computation of the unit step function uses a linear, convex trapezoid as shown in Figure \ref{fig:unitstep-relax}, which is again the tightest over-approximation that we can achieve. The $\actFunOA{j}{\text{U}}{\sSet}$ operation also introduces a new variable $\sPolyVars{m+1}$ and, in this case, four new constraints, which define the trapezoid. The four constraints are as follows: (1) $\sPolyVars{m+1} \geq R_{\text{min}}$, (2) $\sPolyVars{m+1} \leq R_{\text{max}}$, (3) $\sPolyVars{m+1} \leq \frac{R_{\text{max}} - R_{\text{min}}}{val-l}\cdot x_j + \frac{val \cdot R_{\text{min}} - l \cdot R_{\text{max}}}{val - l}$, and (4) $\sPolyVars{m+1} \geq \frac{R_{\text{max}} - R_{\text{min}}}{u - val} \cdot x_j + \frac{u \cdot R_{\text{min}} - val \cdot R_{\text{max}}}{u - val}$. 
As previously, the result of $\actFunOA{j}{\text{U}}{\sSet}$ is a single star which over-approximates the exact resulting star(s). In case the domain of $\sSet$ in dimension $j$ lies completely in either $(-\infty, val]$ or $[val, \infty)$, then the resulting star is either  $\sSet_1 = \tuple{\vectorS{s}_{\text{min}} + \matrixS{M} \sCenter{}, \matrixS{M} \sGeneratorM{}, \sPredicate}$ or $\sSet_2 = \tuple{\vectorS{s}_{\text{max}} + \matrixS{M} \sCenter{}, \matrixS{M} \sGeneratorM{}, \sPredicate}$, respectively.

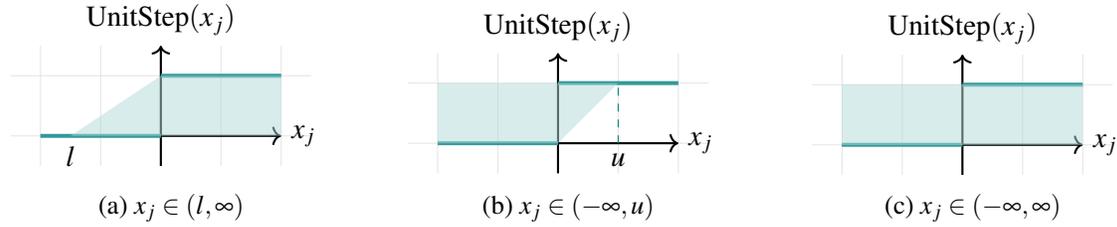
\begin{figure}
    \centering
    \begin{subfigure}{.33\textwidth}
        \centering
        \begin{tikzpicture}[scale=0.8]
    \draw[gray!20,step=1] (-2.5,-0.5) grid (2.5,1.5);
    
    \draw[thick,->] (-2, 0) -- (2, 0) node[right] {$x_j$};
    \draw[thick,->] (0, -0.5) -- (0, 1.5) node[above] {$\text{UnitStep}(x_j)$};
    
    \draw[line width=1.5pt, teal!80] (-2, 0) -- (0, 0); 
    \draw[line width=1.5pt, teal!80] (0, 1) -- (2, 1); 
    
    \fill[teal!30, opacity=0.5] (-1.5, 0) -- (0, 1) -- (2, 1) -- (2, 0);
    
    \node[below] at (-1.5, 0) {$l$};

\end{tikzpicture}
        \caption{$x_j \in (l, \infty)$}
        \label{fig:ustep-1}
    \end{subfigure}%
    \begin{subfigure}{.33\textwidth}
        \centering
        \begin{tikzpicture}[scale=0.8]
    \draw[gray!20,step=1] (-2.5,-0.5) grid (2.5,1.5);
    
    \draw[thick,->] (-2, 0) -- (2, 0) node[right] {$x_j$};
    \draw[thick,->] (0, -0.5) -- (0, 1.5) node[above] {$\text{UnitStep}(x_j)$};
    
    \draw[line width=1.5pt, teal!80] (-2, 0) -- (0, 0); 
    \draw[line width=1.5pt, teal!80] (0, 1) -- (2, 1); 

    \fill[teal!30, opacity=0.5] (-2, 0) -- (-2, 1) -- (1, 1) -- (0, 0);
    
    \draw[teal, dashed] (1,0) -- (1,1);

    \node[below] at (1, 0) {$u$};
\end{tikzpicture}
        \caption{$x_j \in (-\infty, u)$}
        \label{fig:ustep-2}
    \end{subfigure}
    \begin{subfigure}{.33\textwidth}
        \centering
        \begin{tikzpicture}[scale=0.8]
    \draw[gray!20,step=1] (-2.5,-0.5) grid (2.5,1.5);
    
    \draw[thick,->] (-2, 0) -- (2, 0) node[right] {$x_j$};
    \draw[thick,->] (0, -0.5) -- (0, 1.5) node[above] {$\text{UnitStep}(x_j)$};
    
    \draw[line width=1.5pt, teal!80] (-2, 0) -- (0, 0); 
    \draw[line width=1.5pt, teal!80] (0, 1) -- (2, 1); 

    \fill[teal!30, opacity=0.5] (-2, 0) -- (-2, 1) -- (2, 1) -- (2, 0);
\end{tikzpicture}
        \caption{$x_j \in (-\infty, \infty)$}
        \label{fig:ustep-3}
    \end{subfigure}
    \caption{Convex relaxations of the unit stepfunction with three cases of an unbounded input set.}
    \label{fig:ustep-unbounded}
\end{figure}

Finally, in case there is an unbounded input star $\sSet$, again three cases are distinguished, as in case of LeakyReLU. The three cases are as follows: (1) $x_j \in (-\infty, u)$, (2) $x_j \in (l, \infty)$, and (3) $x_j \in (-\infty, \infty)$. The analysis for unbounded input is the same; the only aspect that changes is the introduced constraints. See the corresponding constraints for each case visualized in \autoref{fig:ustep-unbounded}.

\section{Experimental Evaluation}
\label{sec:evaluation}


We implemented our proposed algorithms using the open-source C++ tool HyPro \cite{2017hypro} and evaluated them on four different benchmark families. The ACAS Xu and drone hovering benchmarks contain only ReLU activations while the thermostat and sonar classifier benchmarks use the unit step and hard sigmoid activation functions besides ReLU. Both the exact and the over-approximation approaches are evaluated.
The evaluations were performed on RWTH Aachen University's HPC Cluster \cite{hpc} using Rocky Linux 8 as the operating system. Each execution ran on an individual node equipped with 16GB RAM and two Intel Xeon Platinum 8160 "SkyLake" processors with a total of 16 cores. A 48-hour timeout was set for each experiment.

\subsection{ACAS Xu}
\label{subsec:acas-xu}

\begin{wrapfigure}[15]{r}{0.40\linewidth}
    \centering
    \includegraphics[width=\textwidth]{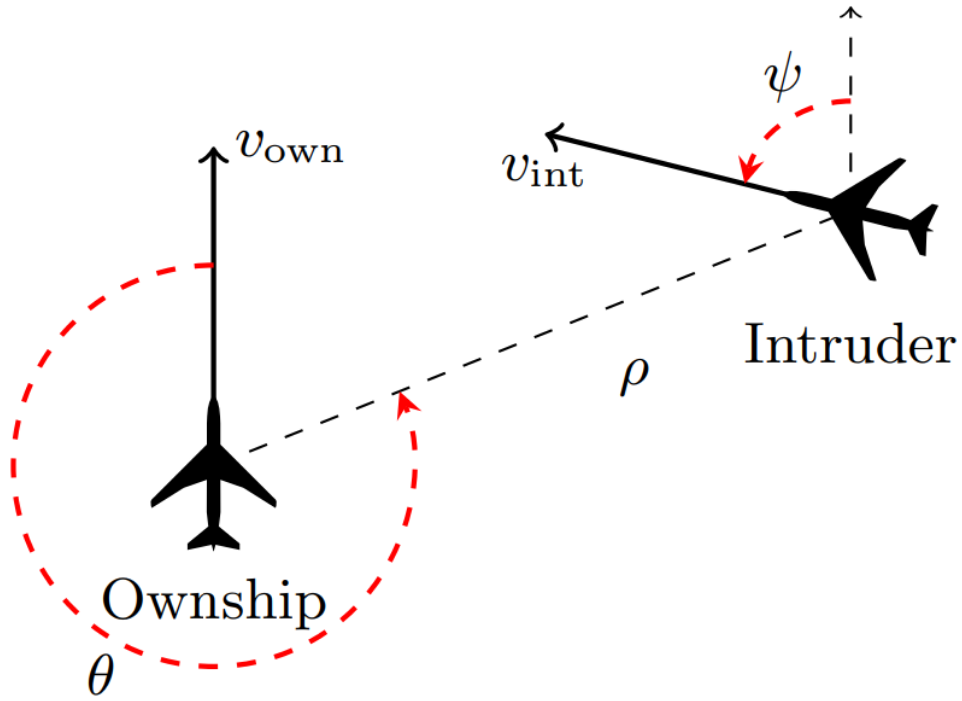}
    \caption{Vertical view of the inputs of ACAS Xu networks. \cite{katz2017reluplex}}
    \label{fig:acas-xu}
\end{wrapfigure}
The Airborne Collision Avoidance System Xu (ACAS Xu) is a mid-air collision avoidance system focusing on unmanned aircrafts.
The ACAS Xu networks (ACAS Xu DNNs) provide advisories for horizontal maneuvers to avoid collisions while minimizing unnecessary alerts.
The ACAS Xu benchmark consists of a set of 45 feedforward neural networks, each with seven fully connected layers, comprising a combined count of 300 neurons.
Each network possesses five inputs (see Figure \ref{fig:acas-xu}) and five outputs. For further information about the ACAS Xu benchmark see \cite{julian_2019,katz2017reluplex}.


For our evaluation, we first compute the reachable set of the networks. Afterward, we check whether the reachable set is fully included in the safe zone. If yes then the FNN is safe, otherwise we can conclude unsafety only for the exact analysis.
We check the \textit{safety verification time} (\textit{VT}) in seconds, using the ten safety properties $\phi_1, \phi_2, \hdots, \phi_{10}$ from \cite{katz2017reluplex}.

\begin{wraptable}[11]{r}{0.40\linewidth}
    \centering
    \begin{tabular}{cS[table-format=5.1]S[table-format=4.1]}
        \toprule
        \hline
        \multirow{2}{*}{prop.} & \multicolumn{1}{|c|}{Exact} & \multicolumn{1}{c}{Overapprox.} \\
        \cline{2-3}
        \multicolumn{1}{c}{} & \multicolumn{1}{|c|}{AVG VT(s)} & \multicolumn{1}{c}{AVG VT(s)} \\
        \hline
        \midrule
        $\phi_{1}$ & $\mathbf{35244.9}$ & 2293.4  \\
        $\phi_{2}$ &          44715.5   & 2316.2  \\
        $\phi_{3}$ &   $\mathbf{279.4}$ &   12.4  \\
        $\phi_{4}$ &    $\mathbf{98.0}$ &   11.4  \\
        \hline
        \bottomrule
        \end{tabular}
    \caption{Average Verification results for properties $\phi_{1}, \phi_{2}, \phi_{3}, \phi_{4}$ in seconds.}
    \label{average}
\end{wraptable}

According to the condensed results, which are shown in Table \ref{average}, we can conclude that the star set approach is able to correctly verify the safety properties. We marked with boldface numbers, where the given property could be verified on all the relevant networks.
In case of exact analysis of $\phi_{2}$, the verification results were correct, but in case of 3 networks, timeout occurred. The over-approximate analysis could verify correctly only a subset of the networks.
We refer to the Appendix \ref{subsec:acas-xu-detailed} of this paper for the detailed results, where we show the reachability result and safety verification times of each property and network combinations. Regarding the running time of the reachability analysis: a meaningful comparison could have been made with the implementation provided in \cite{Tran2021}; however, it is in Matlab and currently, we do not own a Matlab license.

\subsection{Drone Hovering}
\label{subsec:drones}

\begin{table}[h]
    \floatbox[{\capbeside\thisfloatsetup{capbesideposition={right,center},capbesidewidth=5.425cm}}]{table}[\FBwidth]
    {\caption{Evaluation results of the drones benchmark. The network is identified as $ACx$, the lower-right index $y$ shows the tested property. \textit{RT} is the reachable set computation time, and \textit{CT} is the safety checking time, both in seconds.
    \textit{RES} is the safety verification result. True indicates that the given neural network was verified to be safe, with respect to the property. Conversely, False means that the network could not be verified as safe (either because of over-approximation error or due to the network being inherently unsafe). Cells with \textit{(-)} indicate cases where timeout occurs.}\label{dronestable}}
    {
    \begin{tabular}{l|S[table-format=4.1]cS[table-format=2.1]|S[table-format=2.1]cS[table-format=2.1]}
        \toprule
        \hline
        \multirow{2}{*}{$ACx_{y}$} & \multicolumn{3}{c|}{Exact} & \multicolumn{3}{c}{Overapprox.} \\
        \cline{2-7}
        \multicolumn{1}{c|}{} & {RT(s)} & RES & {CT(s)} & {RT(s)} & RES & {CT(s)} \\
        \hline
        \midrule
        $AC1_{1}$ & 61.4 & True & 4.9 & 0.2 & False & 0.0 \\
        $AC1_{2}$ & 0.5 & True & 0.0 & 0.1 & False & 0.0 \\
        $AC2_{1}$ & 462.4 & True & 17.7 & 0.5 & False & 0.0 \\
        $AC2_{2}$ & 0.1 & True & 0.0 & 0.1 & False & 0.0 \\
        $AC3_{1}$ & {-} & - & {-} & 2.9 & False & 0.4 \\
        $AC3_{2}$ & 5.1 & True & 0.1 & 0.2 & False & 0.0 \\
        $AC4_{1}$ & {-} & - & {-} & 8.7 & False & 1.5 \\
        $AC4_{2}$ & 103.0 & True & 5.5 & 0.7 & False & 0.0 \\
        $AC5_{1}$ & 304.8 & True & 26.1 & 0.4 & False & 0.1 \\
        $AC5_{2}$ & 0.1 & False & 0.0 & 0.1 & False & 0.0 \\
        $AC6_{1}$ & 2631.7 & True & 84.4 & 0.7 & False & 0.1 \\
        $AC6_{2}$ & 0.1 & False & 0.0 & 0.1 & False & 0.0 \\
        $AC7_{1}$ & {-} & - & {-} & 2.5 & False & 0.1 \\
        $AC7_{2}$ & 4.1 & True & 0.1 & 0.3 & False & 0.0 \\
        $AC8_{1}$ & {-} & - & {-} & 65.9 & False & 19.8 \\
        $AC8_{2}$ & 0.8 & False & 0.0 & 0.6 & False & 0.0 \\
    \end{tabular}
    }
\end{table}

Autonomous drone control revolves around launching a drone into the air and enabling it to hover at a desired altitude \cite{drones,guidotti2022verification}.
This benchmark consists of eight neural networks. The first four consist of two, and the other four networks consist of three hidden layers, each followed by a ReLU activation function. For further info about the benchmark we refer to \cite{Masara2023}. 
We compute the reachability set of the networks as well as the safety verification using our algorithm and measure the reachable set computation time and safety checking time in seconds.
The networks are verified both with the exact and the over-approximation method. For each neural network we test two properties.
The presented results in Table \ref{dronestable} show, as we would expect, that the over-approximative method is much faster compared to the exact algorithm.
However, the exact method verifies almost every property while the over-approximate approach fails in all cases (though some were inherently unsafe).
This confirms that the over-approximate analysis is more scalable and has a smaller computational cost; however, it sacrifices the completeness of the method.

\subsection{Thermostat Controller}
\label{subsec:thermostat}

This benchmark mentioned in the Master's thesis \cite{thermostat} maintains the room temperature $x$ between $17\degree C$ and $23\degree C$ using a thermostat. It achieves this by activating (mode on) and deactivating (mode off) the heater based on the sensed temperature.
The neural network representing the thermostat's controller is a feedforward neural network with four layers.
The input consists of two neurons that express the temperature $x \in \mathbb{R}$ and the current mode (on or off) as $m \in \{0,1\}$. Furthermore, two hidden layers follow, each with ten neurons. Lastly, using the unit step activation function, the output layer predicts whether the heater should turn on or off, producing the control input $Kh = 15$ or $Kh = 0$, respectively.
We compute the reachable sets to verify the safety of the described NN using our reachability method. 

We tested one safety property of the thermostat controller, the input temperature being between $22\degree$ and $23\degree$, and the thermostat being turned on, i.e., $m = 1$, the expected control output should be the turn off signal.
However, the reachability analysis shows two resulting star sets representing the value 15, meaning that the neural network violates its safety specification.
Therefore, we take those star sets and construct the complete counter input set to falsify the neural network, i.e., prove that it is unsafe. The construction of the complete counter input set works as explained in Theorem 2 of \cite{10.1007/978-3-030-30942-8_39}.

\subsection{Sonar Binary Classifier}
\label{subsec:sonar-classifier}

\begin{table}[t]
    \centering
    \begin{tabular}{l|lr|lr|lr|lr}
        \toprule
        \hline
        \multirow{2}{*}{} & \multicolumn{2}{|c|}{$\delta = 0.01$} & \multicolumn{2}{|c}{$\delta = 0.001$} & \multicolumn{2}{|c}{$\delta = 0.0001$} & \multicolumn{2}{|c}{$\delta = 0.00001$}\\
        \cline{2-9}
        \multicolumn{1}{c|}{} & RT& RES & RT & RES & RT & RES & RT & RES \\
        \hline
        \midrule
        Set 1 & 4359 & False & 783 & True & 263 & True & 102 & True \\
        Set 2 & 206243 & False & 1284 & True & 245 & True & 100 & True \\
        Set 3 & 33945 & True & 3768 & True & 401 & True & 308 & True \\
        Set 4 & 7974 & True & 359 & True & 103 & True & 102 & True \\
    \end{tabular}
    \caption{Local adversarial robustness tests of the exact approach. \textit{RT} is the reachable set computation time in milliseconds.
    \textit{RES} is the safety verification result. True indicates that the neural network correctly classifies the input set, while False means that the network was unable to correctly classify the input set.}
    \label{sonar1}
\end{table}

In this section, we evaluate the robustness of a neural network used for binary classification of a sonar dataset.
This dataset describes sonar chirp returns bouncing off from different objects \cite{sonar}.
It contains $60$ input variables representing the returned beams' strength at different angles.
The verified neural network should be capable of robust binary classification, distinguishing between rocks and metal cylinders.
The neural network consists of one hidden layer with 60 neurons, followed by a ReLU activation and an output layer with a single neuron, followed by the composition of a hard sigmoid and a unit step activation function.
The property we want to verify is the local robustness of the neural network.
A neural network is $\delta$-\textit{locally-robust} at input $x$, if for every $x'$ such that $\norm{x - x'}_\infty \leq \delta$, the network assigns the same output label to $x$ and  $x'$.
Our focus lies in determining the robustness threshold that our verification method can provide for the network (i.e., finding the largest $\delta$ for which the robustness property still holds).
\begin{table}[t]
    \centering
    \begin{tabular}{l|lr|lr|lr|lr}
        \toprule
        \hline
        \multirow{2}{*}{} & \multicolumn{2}{|c|}{$\delta = 0.01$} & \multicolumn{2}{|c}{$\delta = 0.001$} & \multicolumn{2}{|c}{$\delta = 0.0001$} & \multicolumn{2}{|c}{$\delta = 0.00001$}\\
        \cline{2-9}
        \multicolumn{1}{c|}{} & RT& RES & RT & RES & RT & RES & RT & RES \\
        \hline
        \midrule
        Set 1 & 234 & Inconclusive & 205 & True & 163 & True & 103 & True \\
        Set 2 & 396 & Inconclusive & 279 & True & 157 & True & 103 & True \\
        Set 3 & 407 & Inconclusive & 367 & True & 177 & True & 174 & True \\
        Set 4 & 339 & True & 167 & True & 104 & True & 101 & True \\
    \end{tabular}
    \caption{Local adversarial robustness tests of the over-approximate approach. \textit{RT} is the reachable set computation time in milliseconds.
    \textit{RES} is the safety verification result.
    True indicates that the neural network correctly classifies the input set, while False means that the network was unable to correctly classify the input set.
    Additionally, Inconclusive is assigned when the reachability analysis algorithm cannot provide a conclusive answer due to the over-approximation errors.}
    \label{sonar2}
\end{table}

We examine this problem on four inputs of the dataset and four $\delta$ values. The first two inputs should output 1, which means a rock, and the next two 0, which means a metal cylinder.
The True results indicate correct classifications within the robustness threshold ($\forall x'$ being correctly classified), False denotes incorrect predictions. Moreover, in the case of over-approximate analysis, Inconclusive means that the verification result is ambiguous due to over-approximation error.
A comparison between the exact and over-approximative algorithms reveals that the exact algorithm proves network robustness in more cases.
Furthermore, different input sets (meaning a single input and its $\delta$ neighborhood) exhibit varying local robustness. For example, in Table \ref{sonar1}, for Set 2, the optimal $\delta$ value is between 0.01 and 0.001.
Tables \ref{sonar1} and \ref{sonar2} are condensed versions of our experiments, to see the complete results, please check the Appendix \ref{subsec:sonar-detailed} of this paper.

\section{Conclusion}
\label{sec:conclusion}


In this paper, we proposed algorithms for star-based reachability analysis of various activation functions used in feed-forward neural networks.
To ensure generality, we implemented the activation functions with flexibility for adaptation to specific use cases.
We implemented an NNET parser in Hypro to simplify the incorporation of additional benchmarks. The presented evaluation results offer valuable insights into network behavior and safety.
As future work, we plan to integrate further layer types. Consequently, we are planning to integrate a more widely-used standard such as ONNX, for storing and parsing neural network inputs. Moreover, comprehensive experiments and evaluations will offer deeper insights into the performance, accuracy, and limitations of this analysis method when applied to neural networks with other activation functions and layer types, hence, exploring its effectiveness on a more realistic and diverse scale of benchmarks.

We also plan to integrate backpropagation methods using star-sets. Backpropagation is a widely used algorithm for training artificial neural networks, offering numerous advantages in efficient training, scalability, flexibility, and generalization capabilities \cite{backpropagation}. Investigating the compatibility and benefits of incorporating backpropagation with star sets can significantly contribute to the advancement of safe and reliable neural networks.

Finally, we are planning to adapt abstraction refinement techniques (such as CEGAR), to reduce the over-approximation error during the reachable set analysis.
\bigskip

\noindent \textbf{Acknowledgements.} We are grateful to Dario Guidotti, Stefano Demarchi, and Armando Tacchella for generously sharing with us their drone hovering benchmark.
This project has received funding from the European Union's Horizon 2020 programme under the Skłodowska-Curie grant agreement No. 956200.

\nocite{*}
\bibliographystyle{eptcs}
\bibliography{references}

\begin{thebibliography}{10}
\providecommand{\bibitemdeclare}[2]{}
\providecommand{\surnamestart}{}
\providecommand{\surnameend}{}
\providecommand{\urlprefix}{Available at }
\providecommand{\url}[1]{\texttt{#1}}
\providecommand{\href}[2]{\texttt{#2}}
\providecommand{\urlalt}[2]{\href{#1}{#2}}
\providecommand{\doi}[1]{doi:\urlalt{https://doi.org/#1}{#1}}
\providecommand{\eprint}[1]{arXiv:\urlalt{https://arxiv.org/abs/#1}{#1}}
\providecommand{\bibinfo}[2]{#2}

\bibitemdeclare{article}{abiodun2019comprehensive}
\bibitem{abiodun2019comprehensive}
\bibinfo{author}{Oludare~Isaac \surnamestart Abiodun\surnameend},
  \bibinfo{author}{Aman \surnamestart Jantan\surnameend},
  \bibinfo{author}{Abiodun~Esther \surnamestart Omolara\surnameend},
  \bibinfo{author}{Kemi~Victoria \surnamestart Dada\surnameend},
  \bibinfo{author}{Abubakar~Malah \surnamestart Umar\surnameend},
  \bibinfo{author}{Okafor~Uchenwa \surnamestart Linus\surnameend},
  \bibinfo{author}{Humaira \surnamestart Arshad\surnameend},
  \bibinfo{author}{Abdullahi~Aminu \surnamestart Kazaure\surnameend},
  \bibinfo{author}{Usman \surnamestart Gana\surnameend} \&
  \bibinfo{author}{Muhammad~Ubale \surnamestart Kiru\surnameend}
  (\bibinfo{year}{2019}): \emph{\bibinfo{title}{Comprehensive review of
  artificial neural network applications to pattern recognition}}.
\newblock {\slshape \bibinfo{journal}{IEEE Access}} \bibinfo{volume}{7}, pp.
  \bibinfo{pages}{158820--158846}, \doi{10.1109/ACCESS.2019.2945545}.

\bibitemdeclare{article}{anthadupula2021review}
\bibitem{anthadupula2021review}
\bibinfo{author}{Sushma~Priya \surnamestart Anthadupula\surnameend} \&
  \bibinfo{author}{Manasi \surnamestart Gyanchandani\surnameend}
  (\bibinfo{year}{2021}): \emph{\bibinfo{title}{A Review and Performance
  Analysis of Non-Linear Activation Functions in Deep Neural Networks}}.
\newblock {\slshape \bibinfo{journal}{Int. Res. J. Mod. Eng. Technol. Sci}},
  \doi{10.1109/iscid.2009.214.}

\bibitemdeclare{inproceedings}{bak2017serllswi}
\bibitem{bak2017serllswi}
\bibinfo{author}{Stanley \surnamestart Bak\surnameend} \&
  \bibinfo{author}{Parasara~Sridhar \surnamestart Duggirala\surnameend}
  (\bibinfo{year}{2017}): \emph{\bibinfo{title}{Simulation-Equivalent
  Reachability of Large Linear Systems with Inputs}}.
\newblock In \bibinfo{editor}{Rupak \surnamestart Majumdar\surnameend} \&
  \bibinfo{editor}{Viktor \surnamestart Kun{\v{c}}ak\surnameend}, editors:
  {\slshape \bibinfo{booktitle}{Computer Aided Verification}}, pp.
  \bibinfo{pages}{401--420}, \doi{10.1007/978-3-319-63387-9_20}.

\bibitemdeclare{article}{boopathy2019cnn}
\bibitem{boopathy2019cnn}
\bibinfo{author}{Akhilan \surnamestart Boopathy\surnameend},
  \bibinfo{author}{Tsui-Wei \surnamestart Weng\surnameend},
  \bibinfo{author}{Pin-Yu \surnamestart Chen\surnameend},
  \bibinfo{author}{Sijia \surnamestart Liu\surnameend} \& \bibinfo{author}{Luca
  \surnamestart Daniel\surnameend} (\bibinfo{year}{2019}):
  \emph{\bibinfo{title}{CNN-Cert: An Efficient Framework for Certifying
  Robustness of Convolutional Neural Networks}}.
\newblock {\slshape \bibinfo{journal}{Proceedings of the AAAI Conference on
  Artificial Intelligence}} \bibinfo{volume}{33}(\bibinfo{number}{01}), pp.
  \bibinfo{pages}{3240--3247}, \doi{10.1609/aaai.v33i01.33013240}.
\newblock
  \urlprefix\url{https://ojs.aaai.org/index.php/AAAI/article/view/4193}.

\bibitemdeclare{misc}{sonar}
\bibitem{sonar}
\bibinfo{author}{Jason \surnamestart Brownlee\surnameend}
  (\bibinfo{year}{2022}): \emph{\bibinfo{title}{Binary Classification Tutorial
  with the Keras Deep Learning Library}}.
\newblock
  \bibinfo{howpublished}{\url{https://machinelearningmastery.com/binary-classification-tutorial-with-the-keras-deep-learning-library/}}.
\newblock \bibinfo{note}{[Accessed : June 1, 2023]}.

\bibitemdeclare{inproceedings}{cheng2017maximum}
\bibitem{cheng2017maximum}
\bibinfo{author}{Chih-Hong \surnamestart Cheng\surnameend},
  \bibinfo{author}{Georg \surnamestart N{\"u}hrenberg\surnameend} \&
  \bibinfo{author}{Harald \surnamestart Ruess\surnameend}
  (\bibinfo{year}{2017}): \emph{\bibinfo{title}{Maximum resilience of
  artificial neural networks}}.
\newblock In: {\slshape \bibinfo{booktitle}{Automated Technology for
  Verification and Analysis: 15th International Symposium, ATVA 2017, Pune,
  India, October 3--6, 2017, Proceedings 15}},
  \bibinfo{organization}{Springer}, pp. \bibinfo{pages}{251--268},
  \doi{10.1007/978-3-319-68167-2_18}.

\bibitemdeclare{article}{clarke1996formal}
\bibitem{clarke1996formal}
\bibinfo{author}{Edmund~M \surnamestart Clarke\surnameend} \&
  \bibinfo{author}{Jeannette~M \surnamestart Wing\surnameend}
  (\bibinfo{year}{1996}): \emph{\bibinfo{title}{Formal methods: State of the
  art and future directions}}.
\newblock {\slshape \bibinfo{journal}{ACM Computing Surveys (CSUR)}}
  \bibinfo{volume}{28}(\bibinfo{number}{4}), pp. \bibinfo{pages}{626--643},
  \doi{10.1145/242223.242257}.

\bibitemdeclare{techreport}{collobert2004}
\bibitem{collobert2004}
\bibinfo{author}{Ronan \surnamestart Collobert\surnameend}
  (\bibinfo{year}{2004}): \emph{\bibinfo{title}{Large scale machine learning}}.
\newblock \bibinfo{type}{Technical Report},
  \bibinfo{institution}{Universit{\'e} de Paris VI}.

\bibitemdeclare{article}{cubuk2017intriguing}
\bibitem{cubuk2017intriguing}
\bibinfo{author}{Ekin \surnamestart Cubuk\surnameend}, \bibinfo{author}{Barret
  \surnamestart Zoph\surnameend}, \bibinfo{author}{Samuel \surnamestart
  Schoenholz\surnameend} \& \bibinfo{author}{Quoc \surnamestart Le\surnameend}
  (\bibinfo{year}{2017}): \emph{\bibinfo{title}{Intriguing Properties of
  Adversarial Examples}}.

\bibitemdeclare{article}{datta2020surveyonactfunction}
\bibitem{datta2020surveyonactfunction}
\bibinfo{author}{Leonid \surnamestart Datta\surnameend} (\bibinfo{year}{2020}):
  \emph{\bibinfo{title}{A Survey on Activation Functions and their relation
  with Xavier and He Normal Initialization}}.

\bibitemdeclare{misc}{vanishing2023}
\bibitem{vanishing2023}
\bibinfo{author}{\surnamestart Educative\surnameend} (\bibinfo{year}{2023}):
  \emph{\bibinfo{title}{What is the vanishing gradient problem?}}
\newblock
  \bibinfo{howpublished}{\url{https://www.educative.io/answers/what-is-the-vanishing-gradient-problem}}.
\newblock \bibinfo{note}{[Accessed: May 12, 2023]}.

\bibitemdeclare{inproceedings}{ehlers2017formal}
\bibitem{ehlers2017formal}
\bibinfo{author}{Ruediger \surnamestart Ehlers\surnameend}
  (\bibinfo{year}{2017}): \emph{\bibinfo{title}{Formal verification of
  piece-wise linear feed-forward neural networks}}.
\newblock In: {\slshape \bibinfo{booktitle}{Automated Technology for
  Verification and Analysis: 15th International Symposium, ATVA 2017, Pune,
  India, October 3--6, 2017, Proceedings 15}},
  \bibinfo{organization}{Springer}, pp. \bibinfo{pages}{269--286},
  \doi{10.1007/978-3-319-68167-2_19}.

\bibitemdeclare{inproceedings}{erhan2014scalable}
\bibitem{erhan2014scalable}
\bibinfo{author}{Dumitru \surnamestart Erhan\surnameend},
  \bibinfo{author}{Christian \surnamestart Szegedy\surnameend},
  \bibinfo{author}{Alexander \surnamestart Toshev\surnameend} \&
  \bibinfo{author}{Dragomir \surnamestart Anguelov\surnameend}
  (\bibinfo{year}{2014}): \emph{\bibinfo{title}{Scalable Object Detection Using
  Deep Neural Networks}}.
\newblock In: {\slshape \bibinfo{booktitle}{2014 IEEE Conference on Computer
  Vision and Pattern Recognition}}, pp. \bibinfo{pages}{2155--2162},
  \doi{10.1109/CVPR.2014.276}.

\bibitemdeclare{inproceedings}{fromherz2021fast}
\bibitem{fromherz2021fast}
\bibinfo{author}{Aymeric \surnamestart Fromherz\surnameend},
  \bibinfo{author}{Klas \surnamestart Leino\surnameend}, \bibinfo{author}{Matt
  \surnamestart Fredrikson\surnameend}, \bibinfo{author}{Bryan \surnamestart
  Parno\surnameend} \& \bibinfo{author}{Corina \surnamestart
  Pasareanu\surnameend} (\bibinfo{year}{2021}): \emph{\bibinfo{title}{Fast
  Geometric Projections for Local Robustness Certification}}.
\newblock In: {\slshape \bibinfo{booktitle}{International Conference on
  Learning Representations}}.
\newblock \urlprefix\url{https://openreview.net/forum?id=zWy1uxjDdZJ}.

\bibitemdeclare{proceedings}{stepfunction}
\bibitem{stepfunction}
\bibinfo{editor}{Osvaldo \surnamestart Gervasi\surnameend},
  \bibinfo{editor}{Beniamino \surnamestart Murgante\surnameend},
  \bibinfo{editor}{Antonio \surnamestart Lagan{\`{a}}\surnameend},
  \bibinfo{editor}{David \surnamestart Taniar\surnameend},
  \bibinfo{editor}{Youngsong \surnamestart Mun\surnameend} \&
  \bibinfo{editor}{Marina~L. \surnamestart Gavrilova\surnameend}, editors
  (\bibinfo{year}{2008}): \emph{\bibinfo{title}{Computational Science and Its
  Applications - {ICCSA} 2008, International Conference, Perugia, Italy, June
  30 - July 3, 2008, Proceedings, Part {I}}}. {\slshape
  \bibinfo{series}{Lecture Notes in Computer Science}} \bibinfo{volume}{5072},
  \bibinfo{publisher}{Springer}, \doi{10.1007/978-3-540-69839-5}.

\bibitemdeclare{article}{goodfellow2014explaining}
\bibitem{goodfellow2014explaining}
\bibinfo{author}{Ian~J. \surnamestart Goodfellow\surnameend},
  \bibinfo{author}{Jonathon \surnamestart Shlens\surnameend} \&
  \bibinfo{author}{Christian \surnamestart Szegedy\surnameend}
  (\bibinfo{year}{2014}): \emph{\bibinfo{title}{Explaining and Harnessing
  Adversarial Examples}}.
\newblock {\slshape \bibinfo{journal}{CoRR}} \bibinfo{volume}{abs/1412.6572}.
\newblock \urlprefix\url{https://api.semanticscholar.org/CorpusID:6706414}.

\bibitemdeclare{article}{guidotti2022verification}
\bibitem{guidotti2022verification}
\bibinfo{author}{Dario \surnamestart Guidotti\surnameend}
  (\bibinfo{year}{2022}): \emph{\bibinfo{title}{Verification of Neural Networks
  for Safety and Security-critical Domains}}.
\newblock {\slshape \bibinfo{journal}{ISSN 1613-0073 CEUR Workshop
  Proceedings}}.
\newblock \urlprefix\url{https://ceur-ws.org/Vol-3345/paper10_RiCeRCa3.pdf}.

\bibitemdeclare{misc}{drones}
\bibitem{drones}
\bibinfo{author}{Dario \surnamestart Guidotti\surnameend},
  \bibinfo{author}{Stefano \surnamestart Demarchi\surnameend},
  \bibinfo{author}{Luca \surnamestart Pulina\surnameend} \&
  \bibinfo{author}{Armando \surnamestart Tacchella\surnameend}
  (\bibinfo{year}{2022}): \emph{\bibinfo{title}{Evaluating Reachability
  Algorithms for Neural Networks on NeVer2}}.

\bibitemdeclare{inproceedings}{henriksen2021deepsplit}
\bibitem{henriksen2021deepsplit}
\bibinfo{author}{Patrick \surnamestart Henriksen\surnameend} \&
  \bibinfo{author}{Alessio \surnamestart Lomuscio\surnameend}
  (\bibinfo{year}{2021}): \emph{\bibinfo{title}{DEEPSPLIT: An Efficient
  Splitting Method for Neural Network Verification via Indirect Effect
  Analysis.}}
\newblock In: {\slshape \bibinfo{booktitle}{IJCAI}}, pp.
  \bibinfo{pages}{2549--2555}, \doi{10.24963/ijcai.2021/351}.

\bibitemdeclare{book}{hinchey1995applications}
\bibitem{hinchey1995applications}
\bibinfo{author}{Michael \surnamestart Hinchey\surnameend},
  \bibinfo{author}{Jonathan \surnamestart Bowen\surnameend} \&
  \bibinfo{author}{Christopher \surnamestart Rouff\surnameend}
  (\bibinfo{year}{2006}): \emph{\bibinfo{title}{Introduction to Formal
  Methods}}.
\newblock \bibinfo{publisher}{Springer}, \doi{10.1007/1-84628-271-3_2}.

\bibitemdeclare{article}{6296526}
\bibitem{6296526}
\bibinfo{author}{Geoffrey \surnamestart Hinton\surnameend},
  \bibinfo{author}{Li~\surnamestart Deng\surnameend}, \bibinfo{author}{Dong
  \surnamestart Yu\surnameend}, \bibinfo{author}{George~E. \surnamestart
  Dahl\surnameend}, \bibinfo{author}{Abdel-rahman \surnamestart
  Mohamed\surnameend}, \bibinfo{author}{Navdeep \surnamestart
  Jaitly\surnameend}, \bibinfo{author}{Andrew \surnamestart Senior\surnameend},
  \bibinfo{author}{Vincent \surnamestart Vanhoucke\surnameend},
  \bibinfo{author}{Patrick \surnamestart Nguyen\surnameend},
  \bibinfo{author}{Tara~N. \surnamestart Sainath\surnameend} \&
  \bibinfo{author}{Brian \surnamestart Kingsbury\surnameend}
  (\bibinfo{year}{2012}): \emph{\bibinfo{title}{Deep Neural Networks for
  Acoustic Modeling in Speech Recognition: The Shared Views of Four Research
  Groups}}.
\newblock {\slshape \bibinfo{journal}{IEEE Signal Processing Magazine}}
  \bibinfo{volume}{29}(\bibinfo{number}{6}), pp. \bibinfo{pages}{82--97},
  \doi{10.1109/MSP.2012.2205597}.

\bibitemdeclare{article}{hinton2012deep}
\bibitem{hinton2012deep}
\bibinfo{author}{Geoffrey \surnamestart Hinton\surnameend},
  \bibinfo{author}{Li~\surnamestart Deng\surnameend}, \bibinfo{author}{Dong
  \surnamestart Yu\surnameend}, \bibinfo{author}{George~E \surnamestart
  Dahl\surnameend}, \bibinfo{author}{Abdel-rahman \surnamestart
  Mohamed\surnameend}, \bibinfo{author}{Navdeep \surnamestart
  Jaitly\surnameend}, \bibinfo{author}{Andrew \surnamestart Senior\surnameend},
  \bibinfo{author}{Vincent \surnamestart Vanhoucke\surnameend},
  \bibinfo{author}{Patrick \surnamestart Nguyen\surnameend},
  \bibinfo{author}{Tara~N \surnamestart Sainath\surnameend} et~al.
  (\bibinfo{year}{2012}): \emph{\bibinfo{title}{Deep neural networks for
  acoustic modeling in speech recognition: The shared views of four research
  groups}}.
\newblock {\slshape \bibinfo{journal}{IEEE Signal processing magazine}}
  \bibinfo{volume}{29}(\bibinfo{number}{6}), pp. \bibinfo{pages}{82--97},
  \doi{10.1109/MSP.2012.2205597}.

\bibitemdeclare{article}{huang2019reachnn}
\bibitem{huang2019reachnn}
\bibinfo{author}{Chao \surnamestart Huang\surnameend}, \bibinfo{author}{Jiameng
  \surnamestart Fan\surnameend}, \bibinfo{author}{Wenchao \surnamestart
  Li\surnameend}, \bibinfo{author}{Xin \surnamestart Chen\surnameend} \&
  \bibinfo{author}{Qi~\surnamestart Zhu\surnameend} (\bibinfo{year}{2019}):
  \emph{\bibinfo{title}{Reachnn: Reachability analysis of neural-network
  controlled systems}}.
\newblock {\slshape \bibinfo{journal}{ACM Transactions on Embedded Computing
  Systems (TECS)}} \bibinfo{volume}{18}(\bibinfo{number}{5s}), pp.
  \bibinfo{pages}{1--22}, \doi{10.1145/3358228}.

\bibitemdeclare{inproceedings}{huang2017safety}
\bibitem{huang2017safety}
\bibinfo{author}{Xiaowei \surnamestart Huang\surnameend},
  \bibinfo{author}{Marta \surnamestart Kwiatkowska\surnameend},
  \bibinfo{author}{Sen \surnamestart Wang\surnameend} \& \bibinfo{author}{Min
  \surnamestart Wu\surnameend} (\bibinfo{year}{2017}):
  \emph{\bibinfo{title}{Safety verification of deep neural networks}}.
\newblock In: {\slshape \bibinfo{booktitle}{International conference on
  computer aided verification}}, \bibinfo{organization}{Springer}, pp.
  \bibinfo{pages}{3--29}, \doi{10.1007/978-3-319-63387-9_1}.

\bibitemdeclare{mastersthesis}{thermostat}
\bibitem{thermostat}
\bibinfo{author}{Ruoran~Gabriela \surnamestart Jiang\surnameend}
  (\bibinfo{year}{2023}): \emph{\bibinfo{title}{Verifying ai-controlled hybrid
  systems}}.
\newblock \bibinfo{type}{Master's thesis}, \bibinfo{school}{RWTH Aachen
  University}, \bibinfo{address}{Aachen, Germany}.
\newblock \bibinfo{note}{Available at
  \url{https://ths.rwth-aachen.de/wp-content/uploads/sites/4/master_thesis_jiang.pdf}}.

\bibitemdeclare{article}{julian_2019}
\bibitem{julian_2019}
\bibinfo{author}{Kyle~D. \surnamestart Julian\surnameend},
  \bibinfo{author}{Mykel~J. \surnamestart Kochenderfer\surnameend} \&
  \bibinfo{author}{Michael~P. \surnamestart Owen\surnameend}
  (\bibinfo{year}{2019}): \emph{\bibinfo{title}{Deep Neural Network Compression
  for Aircraft Collision Avoidance Systems}}.
\newblock {\slshape \bibinfo{journal}{Journal of Guidance, Control, and
  Dynamics}} \bibinfo{volume}{42}(\bibinfo{number}{3}), pp.
  \bibinfo{pages}{598--608}, \doi{10.2514/1.g003724}.
\newblock \urlprefix\url{https://doi.org/10.2514%2F1.g003724}.

\bibitemdeclare{inproceedings}{katz2017reluplex}
\bibitem{katz2017reluplex}
\bibinfo{author}{Guy \surnamestart Katz\surnameend}, \bibinfo{author}{Clark
  \surnamestart Barrett\surnameend}, \bibinfo{author}{David~L. \surnamestart
  Dill\surnameend}, \bibinfo{author}{Kyle \surnamestart Julian\surnameend} \&
  \bibinfo{author}{Mykel~J. \surnamestart Kochenderfer\surnameend}
  (\bibinfo{year}{2017}): \emph{\bibinfo{title}{Reluplex: An Efficient SMT
  Solver for Verifying Deep Neural Networks}}.
\newblock In \bibinfo{editor}{Rupak \surnamestart Majumdar\surnameend} \&
  \bibinfo{editor}{Viktor \surnamestart Kun{\v{c}}ak\surnameend}, editors:
  {\slshape \bibinfo{booktitle}{Computer Aided Verification}},
  \bibinfo{publisher}{Springer International Publishing},
  \bibinfo{address}{Cham}, pp. \bibinfo{pages}{97--117},
  \doi{10.1007/978-3-319-63387-9_5}.

\bibitemdeclare{inproceedings}{katz2019marabou}
\bibitem{katz2019marabou}
\bibinfo{author}{Guy \surnamestart Katz\surnameend}, \bibinfo{author}{Derek~A
  \surnamestart Huang\surnameend}, \bibinfo{author}{Duligur \surnamestart
  Ibeling\surnameend}, \bibinfo{author}{Kyle \surnamestart Julian\surnameend},
  \bibinfo{author}{Christopher \surnamestart Lazarus\surnameend},
  \bibinfo{author}{Rachel \surnamestart Lim\surnameend}, \bibinfo{author}{Parth
  \surnamestart Shah\surnameend}, \bibinfo{author}{Shantanu \surnamestart
  Thakoor\surnameend}, \bibinfo{author}{Haoze \surnamestart Wu\surnameend},
  \bibinfo{author}{Aleksandar \surnamestart Zelji{\'c}\surnameend} et~al.
  (\bibinfo{year}{2019}): \emph{\bibinfo{title}{The marabou framework for
  verification and analysis of deep neural networks}}.
\newblock In: {\slshape \bibinfo{booktitle}{International Conference on
  Computer Aided Verification}}, \bibinfo{organization}{Springer}, pp.
  \bibinfo{pages}{443--452}, \doi{10.1007/978-3-030-25540-4_26}.

\bibitemdeclare{inproceedings}{KernBueningSinz2022}
\bibitem{KernBueningSinz2022}
\bibinfo{author}{Philipp \surnamestart Kern\surnameend},
  \bibinfo{author}{Marko~Kleine \surnamestart B{\"{u}}ning\surnameend} \&
  \bibinfo{author}{Carsten \surnamestart Sinz\surnameend}
  (\bibinfo{year}{2022}): \emph{\bibinfo{title}{Optimized Symbolic Interval
  Propagation for Neural Network Verification}}.
\newblock In: {\slshape \bibinfo{booktitle}{1st Workshop on Formal Verification
  of Machine Learning ({WFVML} 2022) colocated with {ICML} 2022: International
  Conference on Machine Learning}}.

\bibitemdeclare{misc}{kumar2019fnne}
\bibitem{kumar2019fnne}
\bibinfo{author}{\surnamestart {Kumar, Niranjan}\surnameend}
  (\bibinfo{year}{2019}): \emph{\bibinfo{title}{Deep Learning: Feedforward
  Neural Networks Explained}}.
\newblock
  \bibinfo{howpublished}{\url{https://medium.com/hackernoon/deep-learning-feedforward-neural-networks-explained-\
  c3 4ae3f084f1}}.
\newblock \bibinfo{note}{[Accessed: May 03, 2023]}.

\bibitemdeclare{article}{kuutti2020survey}
\bibitem{kuutti2020survey}
\bibinfo{author}{Sampo \surnamestart Kuutti\surnameend},
  \bibinfo{author}{Richard \surnamestart Bowden\surnameend},
  \bibinfo{author}{Yaochu \surnamestart Jin\surnameend}, \bibinfo{author}{Phil
  \surnamestart Barber\surnameend} \& \bibinfo{author}{Saber \surnamestart
  Fallah\surnameend} (\bibinfo{year}{2021}): \emph{\bibinfo{title}{A Survey of
  Deep Learning Applications to Autonomous Vehicle Control}}.
\newblock {\slshape \bibinfo{journal}{IEEE Transactions on Intelligent
  Transportation Systems}} \bibinfo{volume}{22}(\bibinfo{number}{2}), pp.
  \bibinfo{pages}{712--733}, \doi{10.1109/TITS.2019.2962338}.

\bibitemdeclare{article}{lecun2015deep}
\bibitem{lecun2015deep}
\bibinfo{author}{Yann \surnamestart LeCun\surnameend}, \bibinfo{author}{Yoshua
  \surnamestart Bengio\surnameend} \& \bibinfo{author}{Geoffrey \surnamestart
  Hinton\surnameend} (\bibinfo{year}{2015}): \emph{\bibinfo{title}{Deep
  learning}}.
\newblock {\slshape \bibinfo{journal}{nature}}
  \bibinfo{volume}{521}(\bibinfo{number}{7553}), pp. \bibinfo{pages}{436--444},
  \doi{10.1038/nature14539}.

\bibitemdeclare{article}{lee2015comparing}
\bibitem{lee2015comparing}
\bibinfo{author}{Andy \surnamestart Lee\surnameend} (\bibinfo{year}{2015}):
  \emph{\bibinfo{title}{Comparing deep neural networks and traditional vision
  algorithms in mobile robotics}}.
\newblock {\slshape \bibinfo{journal}{Swarthmore University}}.
\newblock \urlprefix\url{https://api.semanticscholar.org/CorpusID:10011895}.

\bibitemdeclare{article}{liu2021algorithms}
\bibitem{liu2021algorithms}
\bibinfo{author}{Changliu \surnamestart Liu\surnameend}, \bibinfo{author}{Tomer
  \surnamestart Arnon\surnameend}, \bibinfo{author}{Christopher \surnamestart
  Lazarus\surnameend}, \bibinfo{author}{Christopher \surnamestart
  Strong\surnameend}, \bibinfo{author}{Clark \surnamestart Barrett\surnameend}
  \& \bibinfo{author}{Mykel~J. \surnamestart Kochenderfer\surnameend}
  (\bibinfo{year}{2021}): \emph{\bibinfo{title}{Algorithms for Verifying Deep
  Neural Networks}}.
\newblock {\slshape \bibinfo{journal}{Foundations and Trends in Optimization}}
  \bibinfo{volume}{4}(\bibinfo{number}{3-4}), pp. \bibinfo{pages}{244--404},
  \doi{10.1561/2400000035}.
\newblock \urlprefix\url{http://theory.stanford.edu/~barrett/pubs/LAL+21.pdf}.

\bibitemdeclare{article}{LIU201711}
\bibitem{LIU201711}
\bibinfo{author}{Weibo \surnamestart Liu\surnameend}, \bibinfo{author}{Zidong
  \surnamestart Wang\surnameend}, \bibinfo{author}{Xiaohui \surnamestart
  Liu\surnameend}, \bibinfo{author}{Nianyin \surnamestart Zeng\surnameend},
  \bibinfo{author}{Yurong \surnamestart Liu\surnameend} \&
  \bibinfo{author}{Fuad~E. \surnamestart Alsaadi\surnameend}
  (\bibinfo{year}{2017}): \emph{\bibinfo{title}{A survey of deep neural network
  architectures and their applications}}.
\newblock {\slshape \bibinfo{journal}{Neurocomputing}} \bibinfo{volume}{234},
  pp. \bibinfo{pages}{11--26}, \doi{10.1016/j.neucom.2016.12.038}.
\newblock
  \urlprefix\url{https://www.sciencedirect.com/science/article/pii/S0925231216315533}.

\bibitemdeclare{article}{lomuscio2017approach}
\bibitem{lomuscio2017approach}
\bibinfo{author}{Alessio \surnamestart Lomuscio\surnameend} \&
  \bibinfo{author}{Lalit \surnamestart Maganti\surnameend}
  (\bibinfo{year}{2017}): \emph{\bibinfo{title}{An approach to reachability
  analysis for feed-forward ReLU neural networks}}.

\bibitemdeclare{inproceedings}{Maas2013RectifierNI}
\bibitem{Maas2013RectifierNI}
\bibinfo{author}{Andrew~L. \surnamestart Maas\surnameend}
  (\bibinfo{year}{2013}): \emph{\bibinfo{title}{Rectifier Nonlinearities
  Improve Neural Network Acoustic Models}}.
\newblock In: {\slshape \bibinfo{booktitle}{Proceedings of the International
  Conference on Machine Learning}}, pp. \bibinfo{pages}{1--6}.
\newblock
  \urlprefix\url{https://www.semanticscholar.org/paper/Rectifier-Nonlinearities-Improve-Neural-Network-Maas/367f2c63a6f6a10b3b64b8729d601e69337ee3cc}.

\bibitemdeclare{mastersthesis}{Masara2023}
\bibitem{Masara2023}
\bibinfo{author}{Hana \surnamestart Masara\surnameend} (\bibinfo{year}{2023}):
  \emph{\bibinfo{title}{Star Set-based Reachability Analysis of Neural Networks
  with Differing Layers and Activation Functions}}.
\newblock \bibinfo{type}{Bachelor's thesis}, \bibinfo{school}{RWTH Aachen
  University}, \bibinfo{address}{Aachen, Germany}.
\newblock \bibinfo{note}{Available at
  \url{https://ths.rwth-aachen.de/wp-content/uploads/sites/4/Thesis-Hana-Masara.pdf}}.

\bibitemdeclare{misc}{onnxpage}
\bibitem{onnxpage}
\emph{\bibinfo{title}{ONNX}}.
\newblock \bibinfo{howpublished}{\url{https://onnx.ai/}}.
\newblock \bibinfo{note}{[Accessed: May 30, 2023]}.

\bibitemdeclare{misc}{pytorch2021}
\bibitem{pytorch2021}
 (\bibinfo{year}{2021}): \emph{\bibinfo{title}{{PyTorch}:
  torch.nn.Hardsigmoid}}.
\newblock
  \bibinfo{howpublished}{\url{https://pytorch.org/docs/stable/generated/torch.nn.Hardsigmoid.html}}.
\newblock \bibinfo{note}{Accessed : May 16, 2023}.

\bibitemdeclare{inproceedings}{10.1007/978-3-642-14295-6_24}
\bibitem{10.1007/978-3-642-14295-6_24}
\bibinfo{author}{Luca \surnamestart Pulina\surnameend} \&
  \bibinfo{author}{Armando \surnamestart Tacchella\surnameend}
  (\bibinfo{year}{2010}): \emph{\bibinfo{title}{An Abstraction-Refinement
  Approach to Verification of Artificial Neural Networks}}.
\newblock In \bibinfo{editor}{Tayssir \surnamestart Touili\surnameend},
  \bibinfo{editor}{Byron \surnamestart Cook\surnameend} \&
  \bibinfo{editor}{Paul \surnamestart Jackson\surnameend}, editors: {\slshape
  \bibinfo{booktitle}{Computer Aided Verification}},
  \bibinfo{publisher}{Springer Berlin Heidelberg}, \bibinfo{address}{Berlin,
  Heidelberg}, pp. \bibinfo{pages}{243--257},
  \doi{10.1007/978-3-642-14295-6_24}.

\bibitemdeclare{misc}{deadneuron2021}
\bibitem{deadneuron2021}
\bibinfo{author}{Luthfi \surnamestart Ramadhan\surnameend}
  (\bibinfo{year}{2021}): \emph{\bibinfo{title}{Neural Network: The Dead
  Neuron}}.
\newblock
  \bibinfo{howpublished}{\url{https://towardsdatascience.com/neural-network-the-dead-neuron-eaa92e575748}}.
\newblock \bibinfo{note}{[Accessed: May 14, 2023]}.

\bibitemdeclare{article}{rawat2017deep}
\bibitem{rawat2017deep}
\bibinfo{author}{Waseem \surnamestart Rawat\surnameend} \&
  \bibinfo{author}{Zenghui \surnamestart Wang\surnameend}
  (\bibinfo{year}{2017}): \emph{\bibinfo{title}{Deep convolutional neural
  networks for image classification: A comprehensive review}}.
\newblock {\slshape \bibinfo{journal}{Neural computation}}
  \bibinfo{volume}{29}(\bibinfo{number}{9}), pp. \bibinfo{pages}{2352--2449},
  \doi{10.1162/neco_a_00990}.

\bibitemdeclare{inproceedings}{ruan2018reachability}
\bibitem{ruan2018reachability}
\bibinfo{author}{Wenjie \surnamestart Ruan\surnameend},
  \bibinfo{author}{Xiaowei \surnamestart Huang\surnameend} \&
  \bibinfo{author}{Marta \surnamestart Kwiatkowska\surnameend}
  (\bibinfo{year}{2018}): \emph{\bibinfo{title}{Reachability Analysis of Deep
  Neural Networks with Provable Guarantees}}.
\newblock In: {\slshape \bibinfo{booktitle}{Proceedings of the Twenty-Seventh
  International Joint Conference on Artificial Intelligence, {IJCAI-18}}},
  \bibinfo{publisher}{International Joint Conferences on Artificial
  Intelligence Organization}, pp. \bibinfo{pages}{2651--2659},
  \doi{10.24963/ijcai.2018/368}.
\newblock \urlprefix\url{https://doi.org/10.24963/ijcai.2018/368}.

\bibitemdeclare{article}{samek2021explaining}
\bibitem{samek2021explaining}
\bibinfo{author}{Wojciech \surnamestart Samek\surnameend},
  \bibinfo{author}{Gr{\'e}goire \surnamestart Montavon\surnameend},
  \bibinfo{author}{Sebastian \surnamestart Lapuschkin\surnameend},
  \bibinfo{author}{Christopher~J \surnamestart Anders\surnameend} \&
  \bibinfo{author}{Klaus-Robert \surnamestart M{\"u}ller\surnameend}
  (\bibinfo{year}{2021}): \emph{\bibinfo{title}{Explaining deep neural networks
  and beyond: A review of methods and applications}}.
\newblock {\slshape \bibinfo{journal}{Proceedings of the IEEE}}
  \bibinfo{volume}{109}(\bibinfo{number}{3}), pp. \bibinfo{pages}{247--278},
  \doi{10.1109/JPROC.2021.3060483}.

\bibitemdeclare{inproceedings}{2017hypro}
\bibitem{2017hypro}
\bibinfo{author}{Stefan \surnamestart Schupp\surnameend},
  \bibinfo{author}{Erika \surnamestart {\'A}brah{\'a}m\surnameend},
  \bibinfo{author}{Ibtissem \surnamestart Makhlouf\surnameend} \&
  \bibinfo{author}{Stefan \surnamestart Kowalewski\surnameend}
  (\bibinfo{year}{2017}): \emph{\bibinfo{title}{HyPro: A C++ Library of State
  Set Representations for Hybrid Systems Reachability Analysis}}.
\newblock In \bibinfo{editor}{Clark \surnamestart Barrett\surnameend},
  \bibinfo{editor}{Misty \surnamestart Davies\surnameend} \&
  \bibinfo{editor}{Temesghen \surnamestart Kahsai\surnameend}, editors:
  {\slshape \bibinfo{booktitle}{NASA Formal Methods}}, pp.
  \bibinfo{pages}{288--294}, \doi{10.1007/978-3-319-57288-8_20}.

\bibitemdeclare{inproceedings}{schupp2017h}
\bibitem{schupp2017h}
\bibinfo{author}{Stefan \surnamestart Schupp\surnameend},
  \bibinfo{author}{Erika \surnamestart {\'A}brah{\'a}m\surnameend},
  \bibinfo{author}{Ibtissem~Ben \surnamestart Makhlouf\surnameend} \&
  \bibinfo{author}{Stefan \surnamestart Kowalewski\surnameend}
  (\bibinfo{year}{2017}): \emph{\bibinfo{title}{H y P ro: A C++ library of
  state set representations for hybrid systems reachability analysis}}.
\newblock In: {\slshape \bibinfo{booktitle}{NASA Formal Methods Symposium}},
  \bibinfo{organization}{Springer}, pp. \bibinfo{pages}{288--294},
  \doi{10.1007/978-3-319-57288-8_20}.

\bibitemdeclare{phdthesis}{Schupp2019StateSR}
\bibitem{Schupp2019StateSR}
\bibinfo{author}{Stefan \surnamestart Schupp\surnameend},
  \bibinfo{author}{Goran \surnamestart Frehse\surnameend} \&
  \bibinfo{author}{Erika \surnamestart {\'A}brah{\'a}m\surnameend}
  (\bibinfo{year}{2019}): \emph{\bibinfo{title}{State set representations and
  their usage in the reachability analysis of hybrid systems}}.
\newblock Ph.D. thesis, \bibinfo{school}{RWTH Aachen University},
  \doi{10.18154/RWTH-2019-08875}.
\newblock
  \urlprefix\url{https://publications.rwth-aachen.de/record/767529/files/767529.pdf}.

\bibitemdeclare{article}{DBLP:journals/pacmpl/SinghGPV19}
\bibitem{DBLP:journals/pacmpl/SinghGPV19}
\bibinfo{author}{Gagandeep \surnamestart Singh\surnameend},
  \bibinfo{author}{Timon \surnamestart Gehr\surnameend},
  \bibinfo{author}{Markus \surnamestart P{\"{u}}schel\surnameend} \&
  \bibinfo{author}{Martin~T. \surnamestart Vechev\surnameend}
  (\bibinfo{year}{2019}): \emph{\bibinfo{title}{An abstract domain for
  certifying neural networks}}.
\newblock {\slshape \bibinfo{journal}{Proc. {ACM} Program. Lang.}}
  \bibinfo{volume}{3}(\bibinfo{number}{{POPL}}), pp.
  \bibinfo{pages}{41:1--41:30}, \doi{10.1145/3290354}.
\newblock \urlprefix\url{https://doi.org/10.1145/3290354}.

\bibitemdeclare{inproceedings}{sun2021probabilistic}
\bibitem{sun2021probabilistic}
\bibinfo{author}{Bing \surnamestart Sun\surnameend}, \bibinfo{author}{Jun
  \surnamestart Sun\surnameend}, \bibinfo{author}{Ting \surnamestart
  Dai\surnameend} \& \bibinfo{author}{Lijun \surnamestart Zhang\surnameend}
  (\bibinfo{year}{2021}): \emph{\bibinfo{title}{Probabilistic Verification of
  Neural Networks Against Group Fairness}}.
\newblock In: {\slshape \bibinfo{booktitle}{Formal Methods: 24th International
  Symposium, FM 2021, Virtual Event, November 20-26, 2021, Proceedings}},
  \bibinfo{publisher}{Springer-Verlag}, \bibinfo{address}{Berlin, Heidelberg},
  pp. \bibinfo{pages}{83--102}, \doi{10.1007/978-3-030-90870-6_5}.
\newblock \urlprefix\url{https://doi.org/10.1007/978-3-030-90870-6_5}.

\bibitemdeclare{article}{svozil1997intromutlifnn}
\bibitem{svozil1997intromutlifnn}
\bibinfo{author}{Daniel \surnamestart Svozil\surnameend},
  \bibinfo{author}{Vladimir \surnamestart Kvasnicka\surnameend} \&
  \bibinfo{author}{Jiri \surnamestart Pospichal\surnameend}
  (\bibinfo{year}{1997}): \emph{\bibinfo{title}{Introduction to multi-layer
  feed-forward neural networks}}.
\newblock {\slshape \bibinfo{journal}{Chemometrics and Intelligent Laboratory
  Systems}} \bibinfo{volume}{39}(\bibinfo{number}{1}), pp.
  \bibinfo{pages}{43--62}, \doi{10.1016/S0169-7439(97)00061-0}.

\bibitemdeclare{misc}{tensorflowhsigmoid}
\bibitem{tensorflowhsigmoid}
 (\bibinfo{year}{2023}): \emph{\bibinfo{title}{TensorFlow Documentation}}.
\newblock
  \bibinfo{howpublished}{\url{https://www.tensorflow.org/api_docs/python/tf/keras/activations/hard_sigmoid}}.
\newblock \bibinfo{note}{Accessed : May 16, 2023}.

\bibitemdeclare{phdthesis}{tran2020verification}
\bibitem{tran2020verification}
\bibinfo{author}{Dung \surnamestart Tran\surnameend} (\bibinfo{year}{2020}):
  \emph{\bibinfo{title}{Verification of Learning-enabled Cyber-Physical
  Systems}}.
\newblock Ph.D. thesis, \bibinfo{school}{Vanderbilt University Graduate
  School}.
\newblock \urlprefix\url{http://hdl.handle.net/1803/15957}.

\bibitemdeclare{inproceedings}{10.1007/978-3-030-30942-8_39}
\bibitem{10.1007/978-3-030-30942-8_39}
\bibinfo{author}{Hoang-Dung \surnamestart Tran\surnameend},
  \bibinfo{author}{Diago \surnamestart Manzanas~Lopez\surnameend},
  \bibinfo{author}{Patrick \surnamestart Musau\surnameend},
  \bibinfo{author}{Xiaodong \surnamestart Yang\surnameend},
  \bibinfo{author}{Luan~Viet \surnamestart Nguyen\surnameend},
  \bibinfo{author}{Weiming \surnamestart Xiang\surnameend} \&
  \bibinfo{author}{Taylor~T. \surnamestart Johnson\surnameend}
  (\bibinfo{year}{2019}): \emph{\bibinfo{title}{Star-Based Reachability
  Analysis of Deep Neural Networks}}.
\newblock In \bibinfo{editor}{Maurice~H. \surnamestart ter Beek\surnameend},
  \bibinfo{editor}{Annabelle \surnamestart McIver\surnameend} \&
  \bibinfo{editor}{Jos{\'e}~N. \surnamestart Oliveira\surnameend}, editors:
  {\slshape \bibinfo{booktitle}{Formal Methods -- The Next 30 Years}},
  \bibinfo{publisher}{Springer International Publishing},
  \bibinfo{address}{Cham}, pp. \bibinfo{pages}{670--686},
  \doi{10.1007/978-3-030-30942-8_39}.

\bibitemdeclare{inproceedings}{tran2019parallelizable}
\bibitem{tran2019parallelizable}
\bibinfo{author}{Hoang-Dung \surnamestart Tran\surnameend},
  \bibinfo{author}{Patrick \surnamestart Musau\surnameend},
  \bibinfo{author}{Diego~Manzanas \surnamestart Lopez\surnameend},
  \bibinfo{author}{Xiaodong \surnamestart Yang\surnameend},
  \bibinfo{author}{Luan~Viet \surnamestart Nguyen\surnameend},
  \bibinfo{author}{Weiming \surnamestart Xiang\surnameend} \&
  \bibinfo{author}{Taylor~T \surnamestart Johnson\surnameend}
  (\bibinfo{year}{2019}): \emph{\bibinfo{title}{Parallelizable reachability
  analysis algorithms for feed-forward neural networks}}.
\newblock In: {\slshape \bibinfo{booktitle}{2019 IEEE/ACM 7th International
  Conference on Formal Methods in Software Engineering (FormaliSE)}},
  \bibinfo{organization}{IEEE}, pp. \bibinfo{pages}{51--60},
  \doi{10.1109/FormaliSE.2019.00012}.

\bibitemdeclare{article}{tran2021verification}
\bibitem{tran2021verification}
\bibinfo{author}{Hoang-Dung \surnamestart Tran\surnameend},
  \bibinfo{author}{Neelanjana \surnamestart Pal\surnameend},
  \bibinfo{author}{Diego~Manzanas \surnamestart Lopez\surnameend},
  \bibinfo{author}{Patrick \surnamestart Musau\surnameend},
  \bibinfo{author}{Xiaodong \surnamestart Yang\surnameend},
  \bibinfo{author}{Luan~Viet \surnamestart Nguyen\surnameend},
  \bibinfo{author}{Weiming \surnamestart Xiang\surnameend},
  \bibinfo{author}{Stanley \surnamestart Bak\surnameend} \&
  \bibinfo{author}{Taylor~T \surnamestart Johnson\surnameend}
  (\bibinfo{year}{2021}): \emph{\bibinfo{title}{Verification of piecewise deep
  neural networks: a star set approach with zonotope pre-filter}}.
\newblock {\slshape \bibinfo{journal}{Formal Aspects of Computing}}
  \bibinfo{volume}{33}, pp. \bibinfo{pages}{519--545},
  \doi{10.1007/s00165-021-00553-4}.

\bibitemdeclare{article}{Tran2021}
\bibitem{Tran2021}
\bibinfo{author}{Hoang-Dung \surnamestart Tran\surnameend},
  \bibinfo{author}{Neelanjana \surnamestart Pal\surnameend},
  \bibinfo{author}{Diego~Manzanas \surnamestart Lopez\surnameend},
  \bibinfo{author}{Patrick \surnamestart Musau\surnameend},
  \bibinfo{author}{Xiaodong \surnamestart Yang\surnameend},
  \bibinfo{author}{Luan~Viet \surnamestart Nguyen\surnameend},
  \bibinfo{author}{Weiming \surnamestart Xiang\surnameend},
  \bibinfo{author}{Stanley \surnamestart Bak\surnameend} \&
  \bibinfo{author}{Taylor~T. \surnamestart Johnson\surnameend}
  (\bibinfo{year}{2021}): \emph{\bibinfo{title}{Verification of piecewise deep
  neural networks: a star set approach with zonotope pre-filter}}.
\newblock {\slshape \bibinfo{journal}{Formal Aspects of Computing}}
  \bibinfo{volume}{33}(\bibinfo{number}{4}), pp. \bibinfo{pages}{519--545},
  \doi{10.1007/s00165-021-00553-4}.
\newblock \urlprefix\url{https://doi.org/10.1007/s00165-021-00553-4}.

\bibitemdeclare{misc}{hpc}
\bibitem{hpc}
\bibinfo{author}{RWTH \surnamestart University\surnameend}
  (\bibinfo{year}{2023}): \emph{\bibinfo{title}{RWTH High Performance Computing
  (Linux)}}.
\newblock
  \bibinfo{howpublished}{\url{https://help.itc.rwth-aachen.de/service/rhr4fjjutttf/}}.
\newblock \bibinfo{note}{[Accessed : July 24, 2023]}.

\bibitemdeclare{inproceedings}{wang2018efficient}
\bibitem{wang2018efficient}
\bibinfo{author}{Shiqi \surnamestart Wang\surnameend}, \bibinfo{author}{Kexin
  \surnamestart Pei\surnameend}, \bibinfo{author}{Justin \surnamestart
  Whitehouse\surnameend}, \bibinfo{author}{Junfeng \surnamestart
  Yang\surnameend} \& \bibinfo{author}{Suman \surnamestart Jana\surnameend}
  (\bibinfo{year}{2018}): \emph{\bibinfo{title}{Efficient Formal Safety
  Analysis of Neural Networks}}.
\newblock In: {\slshape \bibinfo{booktitle}{Proceedings of the 32nd
  International Conference on Neural Information Processing Systems}},
  \bibinfo{series}{NIPS'18}, \bibinfo{publisher}{Curran Associates Inc.},
  \bibinfo{address}{Red Hook, NY, USA}, pp. \bibinfo{pages}{6369--6379}.

\bibitemdeclare{article}{wing1990specifier}
\bibitem{wing1990specifier}
\bibinfo{author}{Jeannette~M \surnamestart Wing\surnameend}
  (\bibinfo{year}{1990}): \emph{\bibinfo{title}{A specifier's introduction to
  formal methods}}.
\newblock {\slshape \bibinfo{journal}{Computer}}
  \bibinfo{volume}{23}(\bibinfo{number}{9}), pp. \bibinfo{pages}{8--22},
  \doi{10.1109/2.58215}.

\bibitemdeclare{article}{woodcock2009formal}
\bibitem{woodcock2009formal}
\bibinfo{author}{Jim \surnamestart Woodcock\surnameend},
  \bibinfo{author}{Peter~Gorm \surnamestart Larsen\surnameend},
  \bibinfo{author}{Juan \surnamestart Bicarregui\surnameend} \&
  \bibinfo{author}{John \surnamestart Fitzgerald\surnameend}
  (\bibinfo{year}{2009}): \emph{\bibinfo{title}{Formal methods: Practice and
  experience}}.
\newblock {\slshape \bibinfo{journal}{ACM computing surveys (CSUR)}}
  \bibinfo{volume}{41}(\bibinfo{number}{4}), pp. \bibinfo{pages}{1--36},
  \doi{10.1145/1592434.1592436}.

\bibitemdeclare{article}{backpropagation}
\bibitem{backpropagation}
\bibinfo{author}{Logan~G. \surnamestart Wright\surnameend},
  \bibinfo{author}{Tatsuhiro \surnamestart Onodera\surnameend},
  \bibinfo{author}{Martin~M. \surnamestart Stein\surnameend},
  \bibinfo{author}{Tianyu \surnamestart Wang\surnameend},
  \bibinfo{author}{Darren~T. \surnamestart Schachter\surnameend},
  \bibinfo{author}{Zoey \surnamestart Hu\surnameend} \&
  \bibinfo{author}{Peter~L. \surnamestart McMahon\surnameend}
  (\bibinfo{year}{2022}): \emph{\bibinfo{title}{Deep physical neural networks
  trained with backpropagation}}.
\newblock {\slshape \bibinfo{journal}{Nature}}
  \bibinfo{volume}{601}(\bibinfo{number}{7894}), pp. \bibinfo{pages}{549--555},
  \doi{10.1038/s41586-021-04223-6}.
\newblock \urlprefix\url{https://doi.org/10.1038/s41586-021-04223-6}.

\bibitemdeclare{inproceedings}{wu2022efficient}
\bibitem{wu2022efficient}
\bibinfo{author}{Haoze \surnamestart Wu\surnameend},
  \bibinfo{author}{Aleksandar \surnamestart Zelji{\'{c}}\surnameend},
  \bibinfo{author}{Guy \surnamestart Katz\surnameend} \& \bibinfo{author}{Clark
  \surnamestart Barrett\surnameend} (\bibinfo{year}{2022}):
  \emph{\bibinfo{title}{Efficient Neural Network Analysis with
  Sum-of-Infeasibilities}}.
\newblock In \bibinfo{editor}{Dana \surnamestart Fisman\surnameend} \&
  \bibinfo{editor}{Grigore \surnamestart Rosu\surnameend}, editors: {\slshape
  \bibinfo{booktitle}{Tools and Algorithms for the Construction and Analysis of
  Systems}}, \bibinfo{publisher}{Springer International Publishing},
  \bibinfo{address}{Cham}, pp. \bibinfo{pages}{143--163},
  \doi{10.1007/978-3-030-99524-9_8}.

\bibitemdeclare{article}{xiang2018output}
\bibitem{xiang2018output}
\bibinfo{author}{Weiming \surnamestart Xiang\surnameend},
  \bibinfo{author}{Hoang-Dung \surnamestart Tran\surnameend} \&
  \bibinfo{author}{Taylor~T. \surnamestart Johnson\surnameend}
  (\bibinfo{year}{2018}): \emph{\bibinfo{title}{Output Reachable Set Estimation
  and Verification for Multilayer Neural Networks}}.
\newblock {\slshape \bibinfo{journal}{IEEE Transactions on Neural Networks and
  Learning Systems}} \bibinfo{volume}{29}(\bibinfo{number}{11}), pp.
  \bibinfo{pages}{5777--5783}, \doi{10.1109/TNNLS.2018.2808470}.

\bibitemdeclare{inproceedings}{xu2020reluplex}
\bibitem{xu2020reluplex}
\bibinfo{author}{Jin \surnamestart Xu\surnameend}, \bibinfo{author}{Zishan
  \surnamestart Li\surnameend}, \bibinfo{author}{Bowen \surnamestart
  Du\surnameend}, \bibinfo{author}{Miaomiao \surnamestart Zhang\surnameend} \&
  \bibinfo{author}{Jing \surnamestart Liu\surnameend} (\bibinfo{year}{2020}):
  \emph{\bibinfo{title}{Reluplex made more practical: Leaky ReLU}}.
\newblock In: {\slshape \bibinfo{booktitle}{2020 IEEE Symposium on Computers
  and Communications (ISCC)}}, pp. \bibinfo{pages}{1--7},
  \doi{10.1109/ISCC50000.2020.9219587}.

\end{thebibliography}

\newpage
\appendix
\section{Supplementary Material}
\label{app:supp-material}
\subsection{Formal proofs}
\label{subsec:formal-proofs}

\begin{proposition}[Convex polytopes as stars]
    For any $m,p\in\N$, $\matrixS{C} \in \R^{p \times m}$ and $\vectorS{d} \in \R^p$, the convex polyhedron $\polytope \triangleq \{ \sStarVars{}\in \R^{m} \setIf \matrixS{C} \sStarVars{} \leq \vectorS{d} \}$ can be represented by a star.

    \begin{proof}
        It is straightforward to obtain an equivalent starset $\sSet$ of the polytope $\polytope$, using the null vector as center, i.e., $\sCenter{} = \nullVec$, the set of $n$ unit vectors $\vectorS{e}_i$ for the basis, i.e. $\sGeneratorM{} = \stdBasis$ (i.e., the generator matrix $\sGeneratorM{m} = \mathbb{I}_n$), and the predicate $\sPredicate$ in the form of $\sPredWVarVec \equiv \matrixS{C} \sPolyVarsVec \leq \vectorS{d}$.
    \end{proof}
\end{proposition}

\begin{proposition}[Affine transformation]
Assume an $(n,m)$-dimensional star $\sSet \triangleq \tuple{\sCenter{}, \sGeneratorM{}, \sPredicate}$ and let $\Weights{}\in\R^{k\times n}$ and $\biases{}\in\R^k$.
Then the affine transformation $\{ \Weights{} \sStarVars{} + \biases{} \setIf  \sStarVars{} \in [\sSet] \}$ of $[\sSet]$ is represented by
$\bar{\sSet} = \tuple{\bar{\sCenter{}}, \bar{\sGeneratorM{}}, P}$ with $\bar{\sCenter{}} = \Weights{} \sCenter{} + \biases{}$ and $\bar{\sGeneratorM{}}\in\R^{k\times m}$ with columns $\Weights{} \sGenerator{1}{},  \hdots, \Weights{} \sGenerator{m}{}$.

    \begin{proof}
        Using the definition of the resulting star set after applying the affine transformation, we have $\bar{\sSet} = \{ \vectorS{y} \setIf \vectorS{y} = \Weights{} ( \sCenter{} + \sSum{(\sPolyVars{j} \sGenerator{j}{} )} ) + \biases{} \suchthat \sPredWVarVec \}$. That means, $\bar{\sSet}$ is another star, having the center $\bar{\sCenter{}} = \Weights{} \sCenter{} + \biases{}$ and generator vectors $\sGeneratorM{} = \{ \Weights{} \sGenerator{1}{}, \Weights{} \sGenerator{2}{}, \hdots, \Weights{} \sGenerator{2}{} \}$. Note that the predicate does not change during the computation of the affine mapping of a star. 
    \end{proof}
\end{proposition}

\begin{proposition}[Intersection with halfspace]
    Assume an $(n,m)$-dimensional star $\sSet \triangleq \tuple{\sCenter{}, \sGeneratorM{}, \sPredicate}$ and a half-space $\halfspace \triangleq \{ \sStarVars{}\in\R^n \setIf \vectorS{h}^T \sStarVars{} \leq g \}$ with some $\vectorS{h}\in\R^{n}$ and $g\in\R$. Then the intersetion $[\sSet]\cap\halfspace$ is represented by the star $\bar{\sSet}=\tuple{\sCenter{}, \sGeneratorM{}, \sPredicate \cap {\sPredicate}'}$ with 
    $
    {\sPredicate}' = \{\sPolyVarsVec\in\R^m\,|\, (\vectorS{h}^T  \sGeneratorM{m}) \sPolyVarsVec \leq g - \vectorS{h}^T \sCenter{} \}
    $.

\begin{proof}
    The resulting star is $\bar{\sSet} \triangleq \{ \sStarVars{} \setIf \sStarVars{} = \sCenter{} + \sSum{(\sPolyVars{j} \sGenerator{j}{})} \suchthatshort \sPredWVars \wedge \vectorS{h}^T \sStarVars{} \leq g \}$. Since $\sStarVars{} = \sCenter{} + \sSum{(\sPolyVars{j} \sGenerator{j}{})} $, the new constraint can be written as $\vectorS{h}^T (\sCenter{} + \sGeneratorM{m} \mathbf{\alpha}) \leq g$, where $\sPolyVarsVec = \sPolyVarsAsVec$. Consequently, the new predicate is $\sPredicate \cap {\sPredicate}', \; {\sPredicate}'(\sPolyVarsVec) \triangleq (\vectorS{h}^T \sGeneratorM{m}) \sPolyVarsVec \leq g - \vectorS{h}^T \sCenter{}$.
\end{proof}
\end{proposition}

\begin{proposition}[Emptiness checking]
    A star $\sSet \triangleq \tuple{\sCenter{}, \sGeneratorM{}, \sPredicate}$ is empty if and only if $\sPredicate$ is empty. 

\begin{proof}
    It is straightforward to see that only the predicate restricts the elements of a star. In other words, if the predicte does not allow any solution (i.e., it's \textit{empty}), then the star set is empty as well.
\end{proof}
\end{proposition}

\begin{proposition}[Bounding box]
    Assume an $(n,m)$-dimensional star $\sSet \triangleq \tuple{\sCenter{}, \sGeneratorM{}, \sPredicate}$ with $\sCenter{}=(\sCenter{1},\ldots,\sCenter{n})^T$, and let $\sGeneratorM{}_{(i)}$ be the $i^{th}$ row of $\sGeneratorM{m}$. Let furthermore $B=\{(x_1,\ldots,x_n)^T\in\R^n\,|\,\bigwedge_{i=1}^n \textit{lb}_i\leq x_i\leq \textit{ub}_i\}$ with
        $\textit{lb}_i = \sCenter{i} + \underset{\sPredWVarVec}{\text{min}} \; \sGeneratorM{}_{(i)} \sPolyVarsVec $ 
and
        $\textit{ub}_i = \sCenter{i} + \underset{\sPredWVarVec}{\text{max}} \; \sGeneratorM{}_{(i)} \sPolyVarsVec $ for $i=1,\ldots,n$. Then $[\sSet]\subseteq B$.

\begin{proof}
    According to the star set's definition $\sStarVars{i} = \mathbf{c}_i + \sSum{\sPolyVars{j} \sGenerator{j}{i}}$. That is, if we want to find the lower (or upper) bound of $\sStarVars{i}$, we have to find the solution of $\mathit{minimize}_{x \in \sSet} \; \sStarVars{i}$ (or $\mathit{maximize}_{x \in \sSet} \; \sStarVars{i}$, respecitvely). Using the definition of the star set, we get $\sCenter{i} + \mathit{minimize}_{\sPredWVarVec = \top} \; \sGeneratorM{}^{(i)} \sPolyVarsVec$ (or $\sCenter{i} + \mathit{maximize}_{\sPredWVarVec = \top} \; \sGeneratorM{}^{(i)} \sPolyVarsVec$).
\end{proof}

\end{proposition}


\subsection{ACAS Xu Detailed Results}
\label{subsec:acas-xu-detailed}
    
    \begin{table}[H]
        \centering
\begin{tabular}{l|rrrr|rrrrr}
        \toprule
        \hline
        \multirow{2}{*}{$N_{x,y}$} & \multicolumn{4}{|c|}{Exact} & \multicolumn{4}{|c}{Overapproximation} \\
        \cline{2-9}
        \multicolumn{1}{c|}{} & RT(s) & RES & CT(s) & OSS & RT(s) & RES & CT(s) & OSS \\
        \hline
        \midrule
        $N_{2,1}$ & 15837.45 & False & 325.53 & 193197 & 1338.12 & False & 43.21 & 1 \\
        $N_{2,2}$ & 36112.55 & False & 1174.41 & 472257 & 1005.57 & False & 68.81 & 1 \\
        $N_{2,3}$ & 18532.01 & False & 303.91 & 194275 & 962.63 & False & 74.02 & 1 \\
        $N_{2,4}$ & 8254.17 & False & 299.08 & 114155 & 1391.11 & False & 73.80 & 1 \\
        $N_{2,5}$ & 56241.64 & False & 2275.48 & 677510 & 1790.96 & False & 111.03 & 1 \\
        $N_{2,6}$ & 25991.47 & False & 650.43 & 309631 & 2044.11 & False & 45.39 & 1 \\
\bottomrule
\end{tabular}
\caption{Verification results for property $P_2$ on $36$ ACAS Xu networks. \textit{(RT)} is the reachable set computation time, and \textit{(CT)} is the safety checking time, both in seconds.
\textit{(RES)} is the safety verification result. \textit{(OSS)} describes the number of the output star sets.
The cells with \textit{(-)} correspond to networks in which our algorithm was not able to compute the reachability set successfully in less than 48 hours.}
\end{table}

\begin{table}[H]
    \centering
\begin{tabular}{l|rrrr|rrrrr}
    \toprule
    \hline
    \multirow{2}{*}{$N_{x,y}$} & \multicolumn{4}{|c|}{Exact} & \multicolumn{4}{|c}{Overapproximation} \\
    \cline{2-9}
    \multicolumn{1}{c|}{} & RT(s) & RES & CT(s) & OSS & RT(s) & RES & CT(s) & OSS \\
    \hline
    \midrule
        $N_{1,1}$ & 3013.64 & True & 45.71 & 39835 & 823.72 & False & 7.97 & 1 \\
        $N_{1,2}$ & 3575.72 & True & 53.74 & 45648 & 2214.74 & False & 15.48 & 1 \\
        $N_{1,3}$ & 11037.81 & True & 200.90 & 114287 & 3211.44 & False & 16.37 & 1 \\
        $N_{1,4}$ & 13111.44 & True & 267.78 & 154529 & 2915.33 & False & 21.64 & 1 \\
        $N_{1,5}$ & 9756.54 & True & 196.13 & 122297 & 1618.80 & False & 9.97 & 1 \\
        $N_{1,6}$ & 35718.94 & True & 823.34 & 376647 & 1385.90 & False & 13.06 & 1 \\
\bottomrule
    \end{tabular}
    \caption{Verification results for property $P_1$ on $45$ ACAS Xu networks. \textit{RT} is the reachable set computation time, and \textit{CT} is the safety checking time, both in seconds.
    \textit{RES} is the safety verification result. \textit{OSS} describes the number of the output star sets. }
\end{table}

\begin{table}[H]
    \centering
\begin{tabular}{l|rrrr|rrrrr}
    \toprule
    \hline
    \multirow{2}{*}{$N_{x,y}$} & \multicolumn{4}{|c|}{Exact} & \multicolumn{4}{|c}{Overapproximation} \\
    \cline{2-9}
    \multicolumn{1}{c|}{} & RT(s) & RES & CT(s) & OSS & RT(s) & RES & CT(s) & OSS \\
    \hline
    \midrule
    $N_{1,1}$ & 1768.51 & True & 206.94 & 71930 & 33.88 & False & 1.45 & 1 \\
    $N_{1,2}$ & 1647.32 & True & 115.58 & 40273 & 34.65 & False & 4.28 & 1 \\
    $N_{1,3}$ & 442.65 & True & 29.13 & 12444 & 28.54 & False & 1.87 & 1 \\
    $N_{1,4}$ & 224.24 & True & 3.64 & 4346 & 12.95 & True & 0.09 & 1 \\
    $N_{1,5}$ & 246.37 & True & 4.09 & 4820 & 12.65 & True & 0.11 & 1 \\
    $N_{1,6}$ & 65.95 & True & 0.91 & 1281 & 3.79 & True & 0.03 & 1 \\
\bottomrule
    \end{tabular}
    \caption{Verification results for property $P_3$ on $45$ ACAS Xu networks. \textit{RT} is the reachable set computation time, and \textit{CT} is the safety checking time, both in seconds.
    \textit{RES} is the safety verification result. \textit{OSS} describes the number of the output star sets. }
    \end{table}

    \begin{table}[H]
        \centering
    \begin{tabular}{l|rrrr|rrrrr}
        \toprule
        \hline
        \multirow{2}{*}{$N_{x,y}$} & \multicolumn{4}{|c|}{Exact} & \multicolumn{4}{|c}{Overapproximation} \\
        \cline{2-9}
        \multicolumn{1}{c|}{} & RT(s) & RES & CT(s) & OSS & RT(s) & RES & CT(s) & OSS \\
        \hline
        \midrule
        $N_{1,1}$ & 424.83 & True & 29.77 & 19142 & 12.57 & False & 4.78 & 1 \\
        $N_{1,2}$ & 366.72 & True & 23.35 & 13143 & 18.92 & False & 4.37 & 1 \\
        $N_{1,3}$ & 277.95 & True & 13.40 & 9837 & 22.69 & False & 1.40 & 1 \\
        $N_{1,4}$ & 24.99 & True & 1.42 & 1184 & 5.78 & False & 0.67 & 1 \\
        $N_{1,5}$ & 208.17 & True & 6.32 & 6608 & 6.67 & False & 0.39 & 1 \\
        $N_{1,6}$ & 117.16 & True & 3.32 & 4443 & 9.87 & True & 0.92 & 1 \\
\bottomrule
    \end{tabular}
\caption{Verification results for property $P_4$ on $42$ ACAS Xu networks. \textit{RT} is the reachable set computation time, and \textit{CT} is the safety checking time, both in seconds.
\textit{RES} is the safety verification result. \textit{OSS} describes the number of the output star sets. }
\end{table}
    
\begin{table}[H]
    \centering
\begin{tabular}{l|rrrr|rrrrr}
    \toprule
    \hline
    \multirow{2}{*}{$N_{x,y}$} & \multicolumn{4}{|c|}{Exact} & \multicolumn{4}{|c}{Overapproximation} \\
    \cline{2-9}
    \multicolumn{1}{c|}{} & RT(s) & RES & CT(s) & OSS & RT(s) & RES & CT(s) & OSS \\
    \hline
    \midrule
        $N_{1,1}$ & 2933.99 & True & 352.76 & 59734 & 461.70 & False & 8.63 & 1 \\
        \bottomrule
    \end{tabular}
\caption{Verification results for property $P_5$ on $1$ ACAS Xu network. \textit{RT} is the reachable set computation time, and \textit{CT} is the safety checking time, both in seconds.
\textit{RES} is the safety verification result. \textit{OSS} describes the number of the output star sets. }
\end{table}
    
\begin{table}[H]
    \centering
\begin{tabular}{l|rrrr|rrrrr}
    \toprule
    \hline
    \multirow{2}{*}{$N_{x,y}$} & \multicolumn{4}{|c|}{Exact} & \multicolumn{4}{|c}{Overapproximation} \\
    \cline{2-9}
    \multicolumn{1}{c|}{} & RT(s) & RES & CT(s) & OSS & RT(s) & RES & CT(s) & OSS \\
    \hline
    \midrule
        $N_{1,1}$ & 44026.93 & True & 1159.88 & 187775 & 1083.38 & False & 17.91 & 1 \\
        \bottomrule
    \end{tabular}
\caption{Verification results for property $P_6$ on $1$ ACAS Xu network. \textit{RT} is the reachable set computation time, and \textit{CT} is the safety checking time, both in seconds.
\textit{RES} is the safety verification result. \textit{OSS} describes the number of the output star sets. }
\end{table}
    
\begin{table}[H]
    \centering
\begin{tabular}{l|rrrr|rrrrr}
    \toprule
    \hline
    \multirow{2}{*}{$N_{x,y}$} & \multicolumn{4}{|c|}{Exact} & \multicolumn{4}{|c}{Overapproximation} \\
    \cline{2-9}
    \multicolumn{1}{c|}{} & RT(s) & RES & CT(s) & OSS & RT(s) & RES & CT(s) & OSS \\
    \hline
    \midrule
        $N_{11}$ & - & - & - & - & 1520.91 & False & 147.38 & 1 \\
        \bottomrule
    \end{tabular}
\caption{Verification results for property $P_7$ on $1$ ACAS Xu network. \textit{RT} is the reachable set computation time, and \textit{CT} is the safety checking time, both in seconds.
\textit{RES} is the safety verification result. \textit{OSS} describes the number of the output star sets.
The cells with \textit{(-)} correspond to networks in which our algorithm was not able to compute the reachability set successfully in less than 48 hours}
\end{table}
    
\begin{table}[H]
    \centering
\begin{tabular}{l|rrrr|rrrrr}
    \toprule
    \hline
    \multirow{2}{*}{$N_{x,y}$} & \multicolumn{4}{|c|}{Exact} & \multicolumn{4}{|c}{Overapproximation} \\
    \cline{2-9}
    \multicolumn{1}{c|}{} & RT(s) & RES & CT(s) & OSS & RT(s) & RES & CT(s) & OSS \\
    \hline
    \midrule
    $N_{2,9}$ & - & - & - & - & 1560.52 & False & 46.37 & 1 \\
    \bottomrule
\end{tabular}
\caption{Verification results for property $P_8$ on $1$ ACAS Xu network. \textit{RT} is the reachable set computation time, and \textit{CT} is the safety checking time, both in seconds.
\textit{RES} is the safety verification result. \textit{OSS} describes the number of the output star sets.
The cells with \textit{(-)} correspond to networks in which our algorithm was not able to compute the reachability set successfully in less than 48 hours.}
\end{table}
    
\begin{table}[H]
    \centering
\begin{tabular}{l|rrrr|rrrrr}
    \toprule
    \hline
    \multirow{2}{*}{$N_{x,y}$} & \multicolumn{4}{|c|}{Exact} & \multicolumn{4}{|c}{Overapproximation} \\
    \cline{2-9}
    \multicolumn{1}{c|}{} & RT(s) & RES & CT(s) & OSS & RT(s) & RES & CT(s) & OSS \\
    \hline
    \midrule
        $N_{3,3}$ & 33727.84 & False & 458.09 & 338600 & 541.62 & False & 7.83 & 1 \\
        \bottomrule
    \end{tabular}
    \caption{Verification results for property $P_9$ on $1$ ACAS Xu network. \textit{RT} is the reachable set computation time, and \textit{CT} is the safety checking time, both in seconds.
    \textit{RES} is the safety verification result. \textit{OSS} describes the number of the output star sets. }
    \end{table}
    
    \begin{table}[H]
        \centering
    \begin{tabular}{l|rrrr|rrrrr}
        \toprule
        \hline
        \multirow{2}{*}{$N_{x,y}$} & \multicolumn{4}{|c|}{Exact} & \multicolumn{4}{|c}{Overapproximation} \\
        \cline{2-9}
        \multicolumn{1}{c|}{} & RT(s) & RES & CT(s) & OSS & RT(s) & RES & CT(s) & OSS \\
        \hline
        \midrule
        $N_{4,5}$ & 3281.44 & True & 181.59 & 41088 & 1087.41 & False & 17.68 & 1 \\
        \bottomrule
    \end{tabular}
    \caption{Verification results for property $P_{10}$ on $1$ ACAS Xu network. \textit{RT} is the reachable set computation time, and \textit{CT} is the safety checking time, both in seconds.
    \textit{RES} is the safety verification result. \textit{OSS} describes the number of the output star sets. }
    \end{table}


\subsection{Sonar Binary Classifier Detailed Results}
\label{subsec:sonar-detailed}

\begin{center}
    \begin{longtable}{c|lr|lr|lr|lr}
    
    \toprule
        \hline
        \multirow{2}{*}{} & \multicolumn{2}{|c|}{$\delta = 0.01$} & \multicolumn{2}{|c}{$\delta = 0.001$} & \multicolumn{2}{|c}{$\delta = 0.0001$} & \multicolumn{2}{|c}{$\delta = 0.00001$}\\
        \cline{2-9}
        \multicolumn{1}{c|}{} & RT& RES & RT & RES & RT & RES & RT & RES\\
        \hline
        \midrule
    \endfirsthead
    
    \multicolumn{9}{l}%
    {{\tablename\ \thetable{} -- continued from previous page}} \\
    \hline
        \multirow{2}{*}{} & \multicolumn{2}{|c|}{$\delta = 0.01$} & \multicolumn{2}{|c}{$\delta = 0.001$} & \multicolumn{2}{|c}{$\delta = 0.0001$} & \multicolumn{2}{|c}{$\delta = 0.00001$}\\
        \cline{2-9}
        \multicolumn{1}{c|}{} & RT& RES & RT & RES & RT & RES & RT & RES\\
        \hline
        \midrule\endhead
    
    \hline 
    \multicolumn{9}{r}{{Continued on next page}} \\
    \endfoot
    \endlastfoot
    1 & 4359 & False & 783 & True & 263 & True & 102 & True \\
    2 & 1848 & False & 100 & True & 101 & True & 101 & True \\
    3 & 206243 & False & 1284 & True & 245 & True & 100 & True \\
    4 & 49216 & False & 220 & False & 98 & True & 102 & True \\
    5 & 253978 & False & 1248 & True & 99 & True & 123 & True \\
    6 & 6452 & False & 622 & True & 276 & True & 101 & True \\
    7 & 25640 & False & 98 & False & 99 & False & 103 & False \\
    8 & 646937 & False & 18313 & False & 448 & True & 101 & True \\
\end{longtable}
    
\end{center}

\begin{center}
    \begin{longtable}{c|lr|lr|lr|lr}
    
        \toprule
            \hline
            \multirow{2}{*}{} & \multicolumn{2}{|c|}{$\delta = 0.01$} & \multicolumn{2}{|c}{$\delta = 0.001$} & \multicolumn{2}{|c}{$\delta = 0.0001$} & \multicolumn{2}{|c}{$\delta = 0.00001$}\\
            \cline{2-9}
            \multicolumn{1}{c|}{} & RT& RES & RT & RES & RT & RES & RT & RES\\
            \hline
            \midrule
        \endfirsthead
        
        \multicolumn{9}{l}%
        {{\tablename\ \thetable{} -- continued from previous page}} \\
        \hline
            \multirow{2}{*}{} & \multicolumn{2}{|c|}{$\delta = 0.01$} & \multicolumn{2}{|c}{$\delta = 0.001$} & \multicolumn{2}{|c}{$\delta = 0.0001$} & \multicolumn{2}{|c}{$\delta = 0.00001$}\\
            \cline{2-9}
            \multicolumn{1}{c|}{} & RT& RES & RT & RES & RT & RES & RT & RES\\
            \hline
            \midrule\endhead
        
        \hline 
        \multicolumn{9}{r}{{Continued on next page}} \\
        \endfoot
        \endlastfoot
        1 & 234 & Inconclusive & 205 & True & 163 & True & 103 & True \\
        2 & 278 & True & 103 & True & 102 & True & 103 & True \\
        3 & 396 & Inconclusive & 279 & True & 157 & True & 103 & True \\
        4 & 480 & Inconclusive & 128 & True & 101 & True & 104 & True \\
        5 & 391 & Inconclusive & 260 & True & 101 & True & 104 & True \\
        6 & 341 & True & 203 & True & 157 & True & 103 & True \\
        7 & 437 & Inconclusive & 104 & False & 101 & False & 104 & False \\
        8 & 403 & Inconclusive & 371 & False & 181 & True & 106 & True \\
\end{longtable}
\end{center}

\end{document}